\documentclass[journal]{IEEEtran}
\ifCLASSINFOpdf
\else
\fi
\usepackage{bbm}
\usepackage{graphicx}
\usepackage{subfig}
\usepackage{amsmath,amssymb} 
\usepackage{bbm}
\usepackage{bm}
\usepackage{color}
\usepackage{booktabs}
\usepackage{multirow}
\usepackage{rotating}
\usepackage{mathtools}
\usepackage{algorithm}
\usepackage{algorithmic}
\usepackage[normalem]{ulem} 

\newcommand{\added}[1]{\textcolor{black}{#1}}

\begin{document}
%
\title{Discovery of Shared Semantic Spaces for Multi-Scene Video Query and Summarization}
%
%
%

\author{Xun~Xu,
        Timothy~M.Hospedales,
        and~Shaogang~Gong
\thanks{The authors are with the School of Electronic Engineering and Computer Science, Queen Mary University of London, London E1 4NS, UK.(e-mail: \{xun.xu, t.hospedales, s.gong\}@qmul.ac.uk)}}

%
%

\markboth{IEEE TRANSACTIONS ON CIRCUITS AND SYSTEMS FOR VIDEO TECHNOLOGY}%
{Shell \MakeLowercase{\textit{et al.}}: Bare Demo of IEEEtran.cls for Journals}
%



\maketitle

\begin{abstract}

The growing rate of public space CCTV installations has generated a need for 
automated methods for exploiting video surveillance data including scene 
understanding,  query, behaviour annotation and summarization. For this reason, extensive research has 
been performed on surveillance scene understanding and analysis. However,
most studies have considered single scenes, or groups of adjacent scenes. 
The semantic similarity between different but related scenes (e.g., many different traffic scenes of similar layout) is not generally exploited 
to improve any automated surveillance tasks and reduce manual effort.
Exploiting commonality, and sharing any supervised annotations, between different scenes is however challenging due to:  
Some scenes are totally un-related -- and thus any information sharing between them would be detrimental; 
while others may only share a subset of common activities -- and thus information sharing is only useful if it is selective.
Moreover, semantically similar activities which should be modelled together and shared across scenes
may have quite different pixel-level appearance in each scene.
To address these issues we develop a new framework for distributed multiple-scene global understanding 
that  clusters surveillance scenes by their ability to explain each other's behaviours;
and further discovers which subset of activities are shared versus scene-specific within each cluster. 
 We show how to use this structured representation of multiple scenes to improve common surveillance tasks including
scene activity understanding, cross-scene query-by-example, behaviour classification with reduced supervised labelling requirements, and video 
summarization. In each case we demonstrate  how our multi-scene model improves on a collection of standard
 single scene models and a flat model of all scenes.
\end{abstract}


%
\IEEEpeerreviewmaketitle

\section{Introduction}

\IEEEPARstart{T}{he} widespread use of public space CCTV camera systems has generated
unprecedented amounts of data which can easily overwhelm human operators due to the sheer length of the surveillance videos and the large number of surveillance videos captured at different locations concurrently. This has motivated numerous studies into 
automated means to model, understand, and exploit this data.  Some of the key
tasks addressed by automated surveillance video understanding include: 
(i) Behaviour profiling  / scene understanding to reveal what are the typical activities and behaviours 
 in the surveilled space \cite{journal/pami/WangMG09,journal/IJCV/HospedalesLGX2012,journals/pami/HuXFXTM06,journals/ijcv/VaradarajanEO13,conf/cvpr/KuettelBVF10}; 
 (ii) Behaviour query by example, allowing the operator to search for similar  occurrences to a specified example behaviour \cite{journal/pami/WangMG09}; 
 (iii) Supervised learning to classify/annotate activities or behaviours if events of interest  are annotated in a training dataset
 \cite{journal/IJCV/HospedalesLGX2012}; 
 (iv) Summarization to give an operator a semantic overview of a long video in a short period of time \cite{journals/pami/PritchRP08} and 
 (v) Anomaly  detection to highlight to an operator the most unusual
 events in a recording period
 \cite{journal/pami/WangMG09,journal/IJCV/HospedalesLGX2012,journals/pami/HuXFXTM06}. So far, all of these tasks have generally been addressed within a single scene (single video captured by a static camera), or a group of adjacent scenes. 
  
\begin{figure}[!h]
\begin{center}
\includegraphics[width=0.99\linewidth]{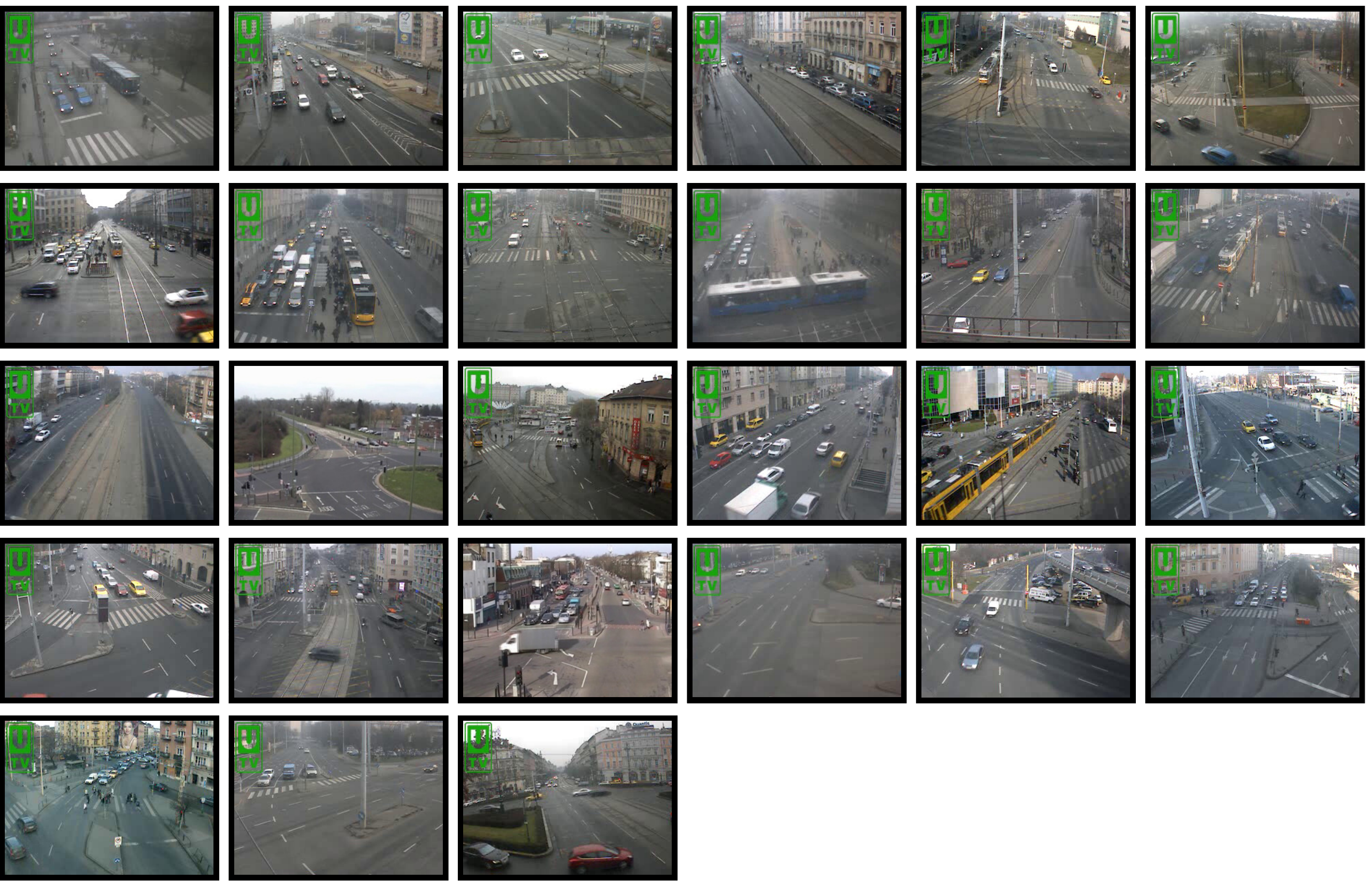}
\caption{An example of multi-camera surveillance network with camera views distributed across different locations.}\label{fig:multiscene_network}
\end{center}
\end{figure}

Compared with single scene recordings, the multi-camera surveillance network (cameras distributed over different locations) is a more realistic scenario in surveillance applications and thus of more interest to end users. An example of a multi-camera surveillance network is given in Fig~\ref{fig:multiscene_network}, where surveillance videos capture mostly traffic scenes with various layouts and motion patterns. In such a multi-scene context, new surveillance tasks arise. For behaviour profiling / scene understanding, human operators would like to see which scenes within the network are semantically similar to each other (e.g. similar scene layout and motion patterns), which activities are in common -- and which are unique -- across a group of scenes, and how activities group into behaviours. Here  activity refers to a spatio-temporally compact motion pattern due to the action of a single or small group of objects (e.g. vehicles making a turn) and
behaviour refers to the interaction between multiple activities within a short temporal segment (e.g. horizontal traffic flow with vehicles going east and west and making a turn). For query-by-example, searching for a specified example behaviour should be carried out not only within scene but also across multiple scenes.
For behaviour classification, annotating training examples in every
scene exhaustively is not scalable. However multi-scene modelling
potentially addresses this by allowing labels to be propagated from
one scene to another. {For summarization, generating a summary
  video for multiple scenes by exploiting cross-scene redundancy can
  provide the user who monitors a set of cameras with an overview of
  all the distinctive behaviours that have occurred in a set of
  scenes. Multi-scene 
  summarisation can reduce the summary length and achieve higher
  compression than single-scene summarization. Combined with
  query-by-example (find more instances of a behaviour in a summary),
  a  flexible exploration of scenes at multiple scales is available.} 

Despite the clear potential benefits of exploiting multi-scene surveillance, it can not be achieved with existing single-scene models \cite{journal/pami/WangMG09,journal/IJCV/HospedalesLGX2012,journals/pami/HuXFXTM06,journals/ijcv/VaradarajanEO13,conf/cvpr/KuettelBVF10}. These approaches learn an independent model for each scene and do not discover corresponding activities or behaviours across scenes even if they share the same semantic meaning. 
This makes any cross-scene reasoning about activities or behaviours impossible. In order to synergistically exploit multiple scenes in surveillance, a multi-scene model with the following capabilities is required: (i) Learning an activity representation that can be shared across scenes; (ii) Model behaviours with the shared representation so they are comparable across scenes and (iii) Generalising surveillance tasks to the multi-scene case, including behaviour profiling/scene understanding, cross-scene query-by-example, cross-scene classification and multi-scene summarization. However this is intrinsically challenging for three reasons:
\begin{enumerate}

\item \textbf{Computing Scene Relatedness}\\
Determining the relatedness of scenes is critical for multi-scene modelling because naive information sharing between insufficiently related scenes can easily result in `negative transfer' \cite{journal/IKDE/PanY2010,Xu:2013:CTS:2510650.2510657}. However, the relatedness of scenes is hard to estimate because the appearance of elements in a scene (e.g. buildings, road surface markings, etc.) is visually diverse, and strongly affected by camera view, making appearance-based similarity measurement unreliable. Similarity measurement based on motion is less prone to visual noise in surveillance applications. However most studies only focus on discovering the similarity in activity level \cite{conf/iccv/KhokharSS11,Xu:2013:CTS:2510650.2510657}. Thus how to measure scene-level relatedness is still an open question.\\

\item \textbf{Selective sharing of information}\\
Large multi-camera surveillance networks covers various types of scenes. Some scenes are totally unrelated which means they convey different semantic meanings to a human. However, more subtly, even between similar scenes, there may be some activities in common and other activities that are unique
to each. Learning a large universal model in this situation is prone to over-fitting due to the high model complexity. Hence a model that discovers (un)relatedness of scenes and selectively shares activities between them is necessary.
\\ 

\item \textbf{Constructing a shared representation}\\
Within related scenes, a shared representation needs to be discovered in order to exploit their similarity for cross-scene query-by-example and multi-scene summarization. Both common and unique activities should be preserved in this process to ensure the ability of discovering not only the commonality but also the distinctiveness between scenes. 
\end{enumerate}

\begin{figure*}[t]
\begin{center}
\includegraphics[width=0.99\linewidth]{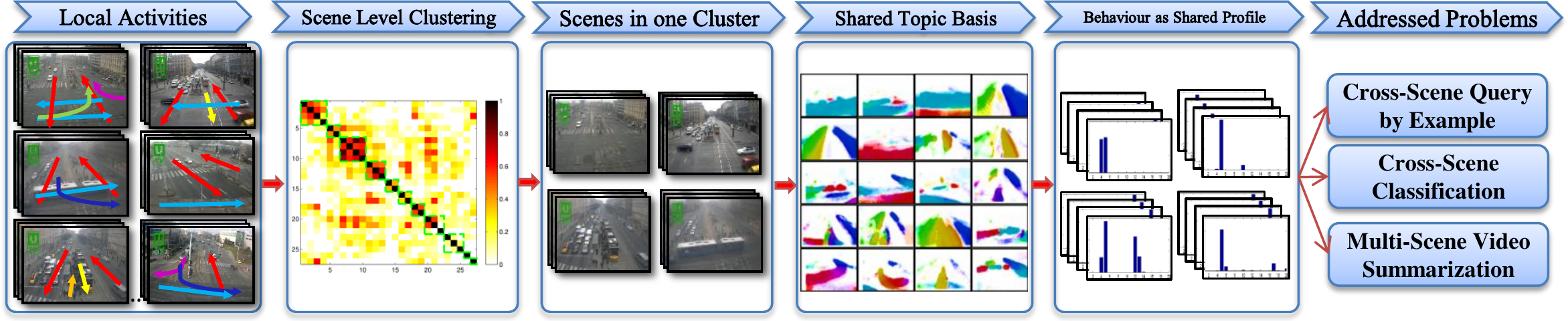}
\end{center}
\caption{An illustration of the proposed framework.}
\label{fig:workflow}
\vspace{-0.5cm\vspace{-0.1cm}}
\end{figure*}

To address these challenges we develop a new framework
illustrated in Fig.~\ref{fig:workflow}. We first learn local representations for each scene separately. Then related scenes are discovered by clustering. A shared semantic representation is constructed to represent activities and behaviours within each group of related scenes. 
Specifically, we first represent each scene with a
low-dimensional `semantic'  (rather than pixel level) representation
through learning a fast unsupervised topic model for
each\footnote{Topics have previously been shown to robustly reveal
  semantic activities from cluttered scenes
  \cite{journal/pami/WangMG09,journal/IJCV/HospedalesLGX2012}.}. Using
a topic-based representation allows us to  
reduce the impact of pixel-noise in discovering activity and scene
similarity. We next group \emph{semantically} related scenes into a scene cluster by
exploiting the correspondence of activities between different scenes. Finally, scenes within each cluster are projected to
a shared representational space by computing a {\em shared activity topic basis} (STB),
shared among all scenes but also allowing each scene to have unique
topics if supported by the data. Behaviours in each scene are represented with the learned STB. 

In addition to profiling for
revealing the multi-scene activity structure across all scenes, 
we use this structured representation to support cross-scene query, label-propagation for classification
and multi-scene summarization. Cross-scene query by example is enabled because
within each cluster, the semantic representation is shared, so an example in one scene can retrieve related examples in every other scene
in the cluster. Behavior classification/annotation in a new scene
\emph{without annotations} is supported because, once associated to a
scene cluster, it can borrow the label-space and classifier from that
cluster. Finally, we define a novel jointly multi-scene
approach to summarization that exploits the shared representation to
 compress redundancy both within and across scenes of each cluster.

\section{Related Work}
\noindent\textbf{Surveillance Scene Understanding}\quad
Scene understanding is a wide area that is too broad to review here. However, some
relevant studies to this work include those based on object tracking
\cite{journals/pami/HuXFXTM06,journals/pami/MorrisT11,journals/ijcv/WangMNG11,conf/iccv/KimLE11},
which model behaviours for example by Hidden Markov Model (HMM)
\cite{journals/pami/HuXFXTM06,journals/pami/MorrisT11}, Gaussian Process \cite{conf/iccv/KimLE11}, clustering \cite{journals/tcsv/PiciarelliMF08} and stochastic context-free grammars \cite{journals/jstsp/FanaswalaK13} and those based
on low-level feature statistics such as optical flow
\cite{DBLP:journals/ijcv/LiGX12,journal/pami/WangMG09,journal/IJCV/HospedalesLGX2012,conf/cvpr/KuettelBVF10}
that often model behaviours by probabilistic topic model (PTM)
\cite{journal/pami/WangMG09,journal/IJCV/HospedalesLGX2012,journals/ijcv/VaradarajanEO13}. The
latter category of approaches are the most related to ours, as we also
built upon PTMs. However, all of these studies operate within-scene
rather than modelling globally distributed scenes and discovering shared activities.\\

\noindent \textbf{Multi-Scene Understanding}\quad
We make an explicit distinction to another line of work that discovers
connections and correlations between multiple overlapping or
non-overlapping scenes connected by a single camera network covering
small areas~\cite{journals/pami/LoyXG12,journals/pami/WangTG10}. This is orthogonal to our
area of interest, which is more similar to multi-task learning \cite{journal/IKDE/PanY2010} - how
to share information between multiple scenes some of which have
semantic similarities, but do not necessarily concurrently surveil topologically
connected zones. 

Fewer approaches have tried to exploit relatedness between scenes without a topological relationship
\cite{conf/iccv/KhokharSS11,Xu:2013:CTS:2510650.2510657}. To recognize
the same activity from another viewpoint, Khorkhar
et al. \cite{conf/iccv/KhokharSS11} proposed a geometric
transformation based method to align two events, represented as
Gaussian mixtures, before computing their similarity. Xu
et al. \cite{Xu:2013:CTS:2510650.2510657} used
a trajectory-based event description and learned motion models from
trajectories observed in a source domain. This model was then used for cross-domain classification
and anomaly detection.

In the context of static image (rather than dynamic scene)
understanding \cite{conf/cvpr/LiPT05,conf/cvpr/LiS009}, studies have
clustered images by appearance similarity. However, this does not
apply directly to surveillance scenes because the background is no
longer stationary nor uniform, e.g. building and road appearance 
are visually salient but can vary significantly between surveillance scenes
at different locations. It is not reliable to relate surveillance
scenes based on appearance - the important cue is activity instead.\\ 
\noindent \textbf{Video Query and Annotation}\quad
Video query has always been an important issue in surveillance
applications. A lot of work has been done on semantic retrieval
\cite{journal/tip/Hu07,journal/pami/WangMG09}. Hu
et al. \cite{journal/tip/Hu07} used trajectories to learn an activity
model and construct semantic indices for video
databases. Wang
et al. \cite{journal/pami/WangMG09} represents video clips as topic
profiles and measures similarity between query and candidate clips as
relative entropy. Retrieved clips are sorted according to the distance
to the query. However none of these techniques take a multi-scene
scenario into consideration, where query examples are selected in
one scene and candidate clips can be retrieved from other scenes at
different locations.

Related to video query, video behaviour annotation/classification has
been addressed in the literature \cite{journal/pami/WangMG09}, also in
terms of video segmentation \cite{XiangGong:PR08}. However, these approaches are typically
domain/scene-specific, which means that \emph{each scene} needs
extensive annotation of training data; where ideally labels should
instead be borrowed from semantically related scenes. Although a
recent study \cite{conf/iccv/KhokharSS11} recognised events across
scenes at the activity level, scene level behaviour classification,
and dealing with a heterogeneous database of scenes is still an open
problem. \\
\noindent\textbf{Video Summarization}\quad
Video summarization has received much attention in the literature in recent years due to the need to digest large quantities of video for efficient review by users. A review  can be found in \cite{journals/tomccap/TruongV07}. There are a variety of approaches to summarization, varying both in how the summary is represented/composed, and how the task is formalised in terms of what type of redundancy should be compressed.

Summaries have been composed by: \emph{static keyframes} that represent the summary as a collection of selected key-frames \cite{journals/prl/AvilaLLA11}, \emph{dynamic skimming} which composes a summary based on a collection of selected clips, and more recently \emph{synopsis}. Synopsis  \cite{conf/iccv/PritchRGP07,journals/pami/PritchRP08} temporally re-orders (spatially non-overlapping) activities from the original video into a temporally compact summary video by shifting activity tubes temporally so they occur more densely. The objective of summarization can be formalised in various ways: to show all foreground activity in the shortest time \cite{conf/iccv/PritchRGP07}, to minimise the reconstruction error between the summary and the original video, to show at least one example of every typical behaviour, or more abstractly to achieve the highest rating in a user study \cite{journals/prl/AvilaLLA11}. 

As the number of scenes grows, multi-view summarization becomes
increasingly important to help operators monitor activities in
numerous scenes. However, multi-view summarization is much less
studied compared to that of single view.  Lou
et al. \cite{conf/mm/LouCL05} adopted multi-view video coding to deal
with multi-view video compression, but did not tackle the more
challenging compression of semantic redundancy. Fu
et al. \cite{journals/tmm/FuGZLSZ10} addressed generating concise
multi-view video summaries by multi-objective optimisation for
generating representative summary clips. Recently, De Leo
et al. \cite{Leo:2014:MVS} proposed a multicamera video summarization
framework which summarizes at the level of activity motif
\cite{conf/bmvc/VaradarajanEO10}. 
Due to the severe occlusion, far-field of view and high density
activities in surveillance videos, none of the existing techniques solve the problem of distributed multi-scene surveillance
video summarization. 

{
In this paper, we pursue video summarization from the
perspective of selecting the smallest set of representative video
clips that still have good \textbf{coverage} of all the behaviours in
the scene(s). Such \emph{multi-scene} summarization compresses redundancy across as well as within scenes.
This corresponds to an application scenario where the
user tasked with monitoring a set of cameras wants an overview
of all the behaviours that  occurred in a set of video streams
during a recording period regardless the source of the video recordings, which typically come from different locations. This perspective on summarization is
attractive because it makes sense of video content indepedent of
location and local context. This offers a more holistic conceptual
summarization in a global context as compared to summarization as
visualisation of a single scene in a local context such as video
synopsis. Interestingly, combined with our query-by-example, we can
take a behaviour of interest shown in the summary as query  to search
for similar behaviours in other scenes. Thus the framework presents
both compact multi-scene summarization and a finer scene-specific zoom-in, capable of compressing semantically equivalent examples no matter what scene they occur in. } \\
\noindent\textbf{Our Contributions}\quad A system based on our framework
can answer questions such as `\emph{show me which scenes are similar
  to this?}' (scene clustering), `\emph{show me which activities are in common and which
  are distinct between these scenes}' (multi-scene profiling)
`\emph{show me all the distinct behaviours in this group of scenes}'
(multi-scene summarization), `\emph{show me other clips from any scene that are
  similar to this nominated example}' (cross-scene query), `\emph{annotate
  this newly provided scene with no-labels}' (cross-scene
classification).  Specifically,  we make the following key contributions:
\begin{enumerate}
\item Introducing the novel and challenging problems of joint multi-scene modelling and analysis.
\item Developing a framework to solve the proposed problem by discovering similarity between activities and scenes, clustering scenes based on semantic similarity and learning a shared representation within scene clusters.
\item We show how to exploit this novel structured multi-scene
model for practical yet challenging tasks of cross-scene query-by-example and behaviour annotation.
\item We further exploit this model to achieve multi-scene video
summarization, achieving compression beyond standard single-scene
approaches.
\item We introduce a large multi-scene surveillance dataset containing 27 distinct views from distributed locations to encourage further investigation into realistic multi-scene visual surveillance applications.
\end{enumerate}

\section{Learning Local Scene Activities}\label{sec:Local}
Given a set of surveillance scenes we first learn local activities
in each individual scene using {\em Latent Dirichlet Allocation}
(LDA) \cite{blei2003}. Although there are more sophisticated
single-scene models 
\cite{journal/pami/WangMG09,journal/IJCV/HospedalesLGX2012,journals/ijcv/VaradarajanEO13},
we use LDA because it is the simplest, most robust, most
generally applicable to a wide variety of scene types, and the fastest
for learning on large scale multi-scene data. However, it could easily be replaced by more elaborate topic models (e.g. HDP \cite{journal/pami/WangMG09}). LDA
generates a set of topics to explain each scene. 
Topics are usually spatially and temporally constrained sub-volumes reflecting the activity of a single or small group of objects.
Following \cite{journal/pami/WangMG09,journal/IJCV/HospedalesLGX2012}, we use
activities to refer to topics and behaviours to refer to
scene-level state defined by the coordinated activities of all scene
participants.
\subsection{Video Clip Representation} \label{subsection:VideoClipRepresentation}
We follow the general approach \cite{journal/pami/WangMG09} to construct visual features for topic models. For each video out of an $M$ scene dataset we first divide the video frame into $N_{a} \times N_{b}$
cells with each cell covering $H \times H$ pixels. Within each cell we
compute optical flow \cite{CLiu:BP}, taking the mean flow as the
motion vector in that cell.  Then we quantize motion vector into
$N_{m}$ fixed directions. {Note, stationary foreground
  objects can be readily added as another cell state as described in
  \cite{journal/IJCV/HospedalesLGX2012,VaradarajanO_ICCV09}.} Therefore 
a codebook \textbf{V} of size $N_{v}=N_{a}\times N_{b} \times N_{m}$
is generated by mapping motion vectors to discrete visual words (from $1$ to $N_{v}$). $N_{d}$ visual documents
$\mathbf{X}=\{\mathbf{x}_{j}\}_{j=1}^{N_{d}}$ are then constructed by
segmenting the video into non-overlapping clips of fixed
length, where each clip $\mathbf{x}_j=\{x_{ij}\}_{i=1}^{N_{j}}$ has $N_{j}$ visual words $x_{ij}$. Clip and document are used interchangeably here with both indicating visual words accumulated in a temporal segment.

\subsection{Learning Local Activities with Topic Model}
\begin{figure}[htb]
\begin{center}
\includegraphics[width=0.75\linewidth]{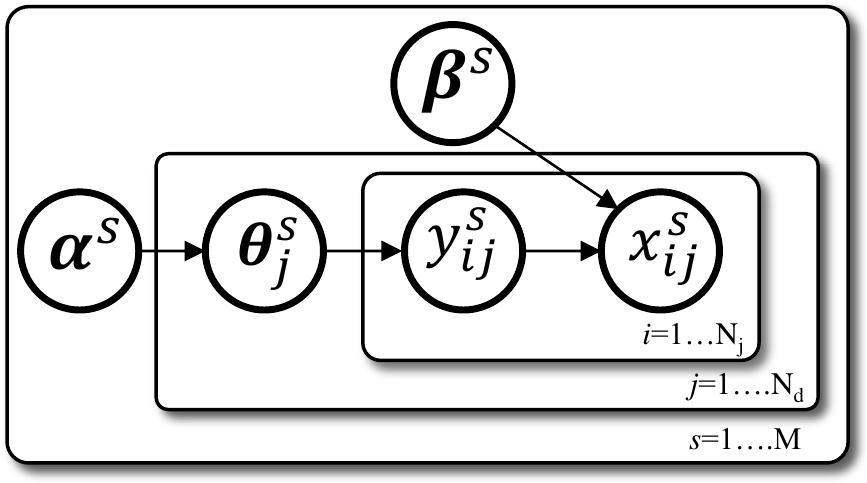}
\end{center}
\caption{Graphical model for Latent Dirichlet Allocation.}
\label{Fig:LDA}
\end{figure}

Learning LDA for scene $s$ discovers the dynamic `appearance' of $k=1\dots K$ typical topics/activities\footnote{In text analysis, a topic refers to a group of co-occurring words in a document. Activity refers to a motion pattern, which defines the group of co-occurring visual words in a video clip. They are used interchangeably in the following text.} (multinomial parameter $\bm\beta_k^s$), and explains each visual word $x_{ij}^s$ in each clip $\mathbf{x}_j^s$  by a latent topic $y_{ij}^s$ specifying which activity generated it, as shown in Fig. \ref{Fig:LDA}. The topic selection $y_{ij}^s$ is drawn from multinomial mixture of topics parametrized by $\bm\theta_j^s$ which is further governed by a Dirchelet distribution with parameter $\bm\alpha^s$. In scene $s$ the joint probability of $N_d$ visual documents $\mathbf{X}^s=\{\mathbf{x}^s_{j}\}_{j=1}^{N_{d}}$, topic selection $\mathbf{Y}^s=\{\mathbf{y}_{j}^s\}_{j=1}^{N_d}$ and topic mixture $\bm\theta^s=\{\bm\theta_j^s\}_{j=1}^{N_d}$ given hyperparameters $\bm\alpha^{s}$ and $\bm\beta^{s}$ is: 

\vspace{-0.2cm}
\begin{equation}
\resizebox{.6\hsize}{!}
{$
\begin{multlined}
p(\bm\theta^s,\mathbf{Y}^s,\mathbf{X}^s \mid \bm\alpha^{s},\bm\beta^{s})=\prod_{j=1}^{N_d} p(\bm\theta_j^s \mid \bm\alpha^{s})
\cdot \\
\prod_{i=1}^{N_j}p(y_{ij}^s\mid \bm\theta_j)p(x_{ij}^s\mid y_{ij}^s,\bm\beta^{s})
\end{multlined}
$}
\end{equation}
\vspace{-0.2cm}

\subsubsection{Model Inference}
Exact inference in LDA is intractable due to the coupling between $\bm\theta$ and $\bm\beta$ \cite{blei2003}. Variational inference approximates a lower bound of log likelihood by introducing variational parameters $\bm\gamma$ and $\bm\phi$. Dirichlet parameter $\bm{\gamma}_j$ is a clip-level topic profile and specifies the mixture ratio of each activity $\bm\beta_k$ in a clip $\mathbf{x}_{j}$. Thus, each  video clip is represented as a mixture of activities ($\bm\gamma_j$). The variational EM procedure for LDA is given in Algorithm~\ref{Alg:ModelInference} where $\mathbbm{1}(\cdot)$ is an indicator function and $\Psi(\cdot)$ is the first derivative of the $\log\Gamma$ function. For
efficiency, we apply the  sparse updates identified in
\cite{fu2013latentAttrib} for an order of magnitude speed increase.

\begin{algorithm}
\caption{Topic model learning for a single scene}\label{Alg:ModelInference}
\begin{algorithmic}
\STATE initialize $\bm\alpha_k=1$
\STATE initialize $\bm\beta=random(N_v,K)$
\STATE initialize $\bm\phi_{ijk}=1/K$
\REPEAT
\STATE \textit{\textbf{E-Step:}}
\FOR{$j=1 \to N_d$}
	\FOR{$k=1 \to K$}
		\STATE $\bm\gamma_{jk}=\bm\alpha_k+\sum_{i=1}^{N_j}\bm\phi_{ijk}$
		\FOR{$i=1 \to N_j$}
			\STATE $\bm\phi_{ijk}=\bm\beta_{x_{ij}k}\exp(\Psi(\bm\gamma_{jk}))$
		\ENDFOR
	\ENDFOR
\ENDFOR
\STATE \textit{\textbf{M-Step:}}
\FOR{$v=1 \to N_v$}
	\FOR{$k=1 \to K$}
		\STATE $\bm\beta_{vk}=\sum_{j=1}^{N_d}\sum_{i=1}^{N_j}\bm\phi_{ijk}\mathbbm{1}(x_{ij}=v)$	
	\ENDFOR
\ENDFOR
\UNTIL{Converge}
\end{algorithmic}
\end{algorithm}

\begin{figure*}[!ht]
\begin{center} 
\includegraphics[width=0.8\linewidth]{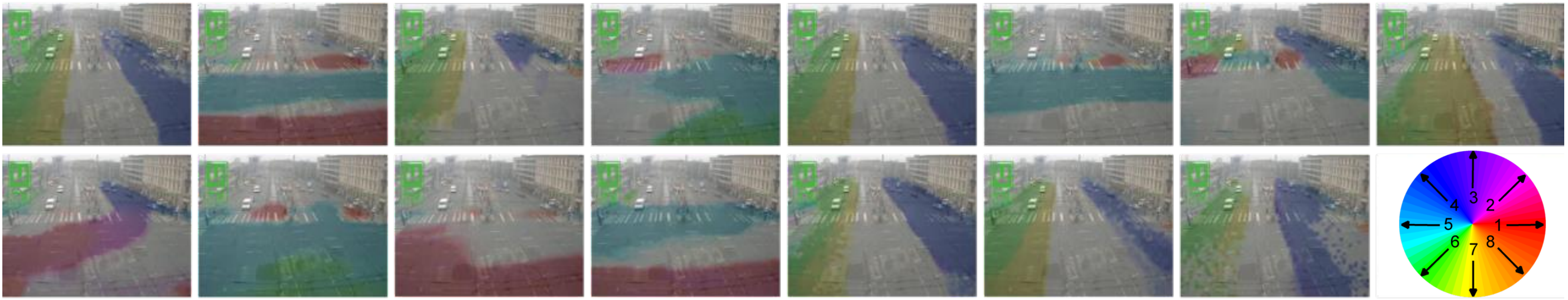}
\end{center}
\caption{Locally learned activities/topics in an example scene. The optical flow is quantized into $N_{m}=8$ directions as shown in the colorwheel.}\label{fig:SceneTopics}
\end{figure*}

After learning all $s=1\dots M$ scenes, every clip $\mathbf{x}_{j}^{s}$
is now represented as a topic profile $\bm\gamma_{j}^s$; and each scene
is now represented by its constituent activities $\bm\beta_{k}^{s}$ (Fig.~\ref{fig:SceneTopics}).

\section{Multi-Layer Activity and Scene Clustering}\label{sec:sceneActCluster}
\begin{figure*}[t]
\begin{center}
\includegraphics[width=0.8\linewidth]{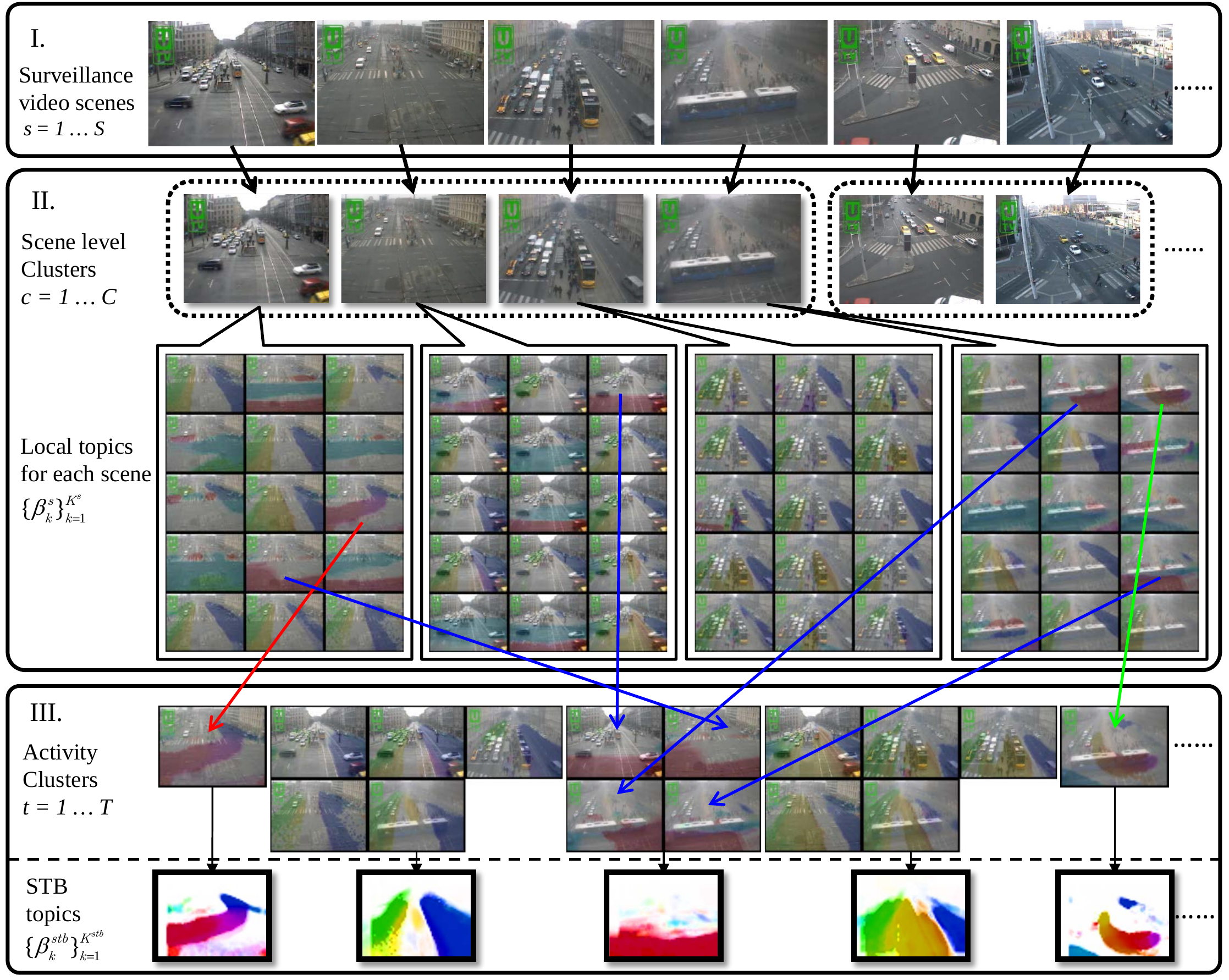}
\end{center}
\caption{An illustration of multi-layer clustering of scenes and activities. Block I (Top) illustrates the original surveillance video scenes. Block II (middle) illustrates (i) related scenes are grouped into clusters (indicated by green dashed boxes) and (ii) the local topics/activities learned in each scene. Block III (bottom) illustrates (i) local topics are furthered grouped into activity clusters (color lines indicate some examples) and (ii) activity clusters are merged to construct a \textit{shared activity topic basis (STB)}.}
\label{Fig:MultilayerClustering}
\vspace{-0.5cm}
\end{figure*}

We next address how to discover related scenes and learn shared topics/activities across scenes. 
This multi-layer process is illustrated in Fig.~\ref{Fig:MultilayerClustering} for two typical clusters 3 \& 7: At the scene level we group related scenes according to activity correspondence (Section~\ref{subsection:SceneLevelClustering}); within each scene cluster we further compute a {\em shared activity topic basis} so that all activities within that cluster are expressed in terms of the same set of topics (Section~\ref{subsection:LearningSTB}).

\subsection{Scene Level Clustering}\label{subsection:SceneLevelClustering}

In order to group related scenes, we first need to define a relatedness metric. Related scenes  should have more common activities so that the model learned from them is compact. So we assume the scenes with semantically similar activities are more likely to be mutually related. We thus define the relatedness between two (aligned) scenes $a$ and $b$, by the correspondence of their semantic activities.

\paragraph{\textbf{Alignment}} Comparing scenes directly suffers from cross-scene variance due to view angle. 
To reduce this cross-scene variance we first align two scenes with a geometrical transformation including scaling $t_{s}$ and translation $[t_{x},t_{y}]$. Although this is not a strong transform it is valid in the typical case that a camera is installed upright, and with surveillance cameras there are classic views which can be simply aligned by scaling and translation. To achieve this, we first denote the transform matrix for normalizing visual words in each scene $a$ and $b$ to the origin as $\mathbf{T}^{a}_{norm}$ and $\mathbf{T}^{b}_{norm}$ defined as Eq.~(\ref{eq:Transform}). Scaling ($t^a_s$) and translation ($t_{x},t_{y}$) parameters are estimated by Eq.~(\ref{eq:TransformParam}).

\begin{equation}\label{eq:Transform}
\resizebox{.4\hsize}{!}
{$
\mathbf{T}^{a}_{norm} = 
\begin{bmatrix}
t^a_s & 0 & t_{x}^{a}\\
0 & t^a_s & t_{y}^{a}\\
0 & 0 & 1
\end{bmatrix}
$}
\end{equation}

\begin{equation}\label{eq:TransformParam}
\resizebox{.7\hsize}{!}
{$
\begin{split}
&center=\frac{1}{N_{d}\cdot N_{j}}\sum_{j=1}^{N_{d}}\sum_{i=1}^{N_{j}}x_{ij}^{a}
 \text{,} \qquad \\
&t^a_s=\frac{N_{d}\cdot N_{j}}{\sum_{j=1}^{N_{d}}\sum_{i=1}^{N_{j}}\|x_{ij}^{a}-center\|_{2}}\text{,} \qquad \\
&\begin{bmatrix}
t_{x}^{a}\\t_{y}^{a}
\end{bmatrix}=
-t^{a}_s \cdot center \qquad \\
\end{split}$}
\end{equation}

\noindent Two scenes can thus be aligned by transforming data from ${a}$ to ${b}$ via $\mathbf{T}^{{a}2{b}} = \mathbf{T}^{b -1}_{norm} \cdot \mathbf{T}^{a}_{norm}$. We then denote $k$th topic in scene $a$ as $\bm\beta_{k}^{a}$. So any topic $k$ in $a$ can be aligned for comparison with those in $b$ by $\mathbf{T}^{{a}2{b}}$.

 { We denote the topic transformation procedure as
   $\bm\beta^{\prime}=\mathbb{H}(\bm\beta;\mathbf{T})$. This transformation is
   applied to topics in a similar way as image transform. That is, given that
   $\bm\beta$ is a $N_{a}\times N_{b} \times N_{m}$ matrix and a
   transform matrix $\mathbf{T}$ is defined as Eq~(\ref{eq:Transform}),
   we first estimate the size $N^{\prime}_a \times N^{\prime}_b \times
   N_m$ of transformed topic $\bm\beta^{\prime}$ by $N^{\prime}_a \!=\!
   N_a\times t_{s}$ and $N^{\prime}_b\!=\! N_b\times t_{s}$. To obtain the
   value for each element/pixel of
   $\bm\beta^{\prime}(x^{\prime},y^{\prime},d^{\prime})$, we trace back
   to the position $[x,y,d]$ in the original topic
   $\bm\beta$. \added{If we only consider scaling and
     translation, direction $d$ is then unchanged throughout the
     procedure i.e. $d^{\prime}=d$. 
Therefore, $x$ and $y$ are determined by:
\begin{equation}\label{eq:Determine_x_y}
\begin{bmatrix}
x & y & 1
\end{bmatrix}=
\begin{bmatrix}
x^{\prime}&
y^{\prime}&
1 
\end{bmatrix}
\cdot
(\mathbf{T}^{-1})^{T}
\end{equation}
}
\vspace{-0.2cm}

{
In most cases, $x$ and $y$ are not discrete values because
  of the matrix multiplication. In order to obtain the value for
  $\bm\beta(x^{\prime},y^{\prime},d^{\prime})$, we perform
  interpolation, i.e. we use the values of adjacent pixels surrounding
  $[x,y,d]$ to determine the value of
  $\bm\beta(x^{\prime},y^{\prime},d^{\prime})$. This interpolation is
  only related to spatial values in a single layer, i.e. $d$ is fixed,
  and we only use the adjacent pixels by varying $x$ and $y$. A number
  of standard interpolation techniques can be used for this task
  including linear, bilinear and bicubic interpolations and we use bicubic interpolation here. After
  interpolation, we compute the exact value for each element/pixel
  $\bm\beta(x^{\prime},y^{\prime},d^{\prime})$. Due to that this
  transformation involves translation, the transformed topic
  $\bm\beta^\prime$ may extend out of the topic boundary, a $N_{a}\times
  N_{b}$ rectangle, defined by the original topic $\bm\beta$. To ensure
  all topics being comparable with the same codebook size, we only keep the
  part of $\bm\beta^\prime$ that lies within the $N_{a}\times N_{b}$
  rectangle defined by the original topic $\bm\beta$. After the above
  procedure, the transformed topic  $\bm\beta^{\prime}$ has the same
  size as the original  $\bm\beta$, $N_{a}\times N_{b} \times
  N_m$. Finally, we normalise the transformed topic
  $\bm\beta^{\prime}$ to obtain a multinomial distribution, as follows:
\begin{equation}\label{eq:BetaNormalize}
\resizebox{.65\hsize}{!}
{$
\bm\beta^{\prime}=\frac{\bm\beta^{\prime}}{\underset{x=1\cdots N_{a}}{\sum}\underset{y=1\cdots N_{b}}{\sum}\underset{d=1\cdots N_{m}}{\sum} \bm\beta^{\prime}(x,y,d)}$}
\end{equation}
}
\paragraph{\textbf{Affinity and clustering}}\quad
Given the scene alignment above, we define the relatedness between
scenes $a$ and $b$ by the percentage of corresponding
topic pairs. More specifically, given $K^{a}$ local topics
$\{\bm\beta_{k^a}^{a}\}_{k_a=1}^{K^a}$ in scene $a$ and $K^{b}$ local topics
$\{\bm\beta_{k^b}^{b}\}_{k_b=1}^{K^b}$ in scene $b$, the distance between topic
$\bm\beta_{k^a}^{a}$ and topic $\bm\beta_{k^b}^{b}$ is defined as
$\mathcal{D}_{\mathbf{KL}}$ in Eq.~(\ref{eq:SymKLDivergence}): 

\begin{equation}\label{eq:SymKLDivergence}
\resizebox{.8\hsize}{!}
{$
\begin{multlined}
\mathcal{D}_{\mathbf{KL}}(\bm\beta_{k^a}^{a},\bm\beta_{k^b}^{b}) = \frac{1}{2}( \mathbf{KL}(\bm\beta_{k^a}^{a2b}\mid \mid \bm\beta_{k^b}^{b}) + 
\mathbf{KL}({\bm\beta_{k^b}^{b2a} \mid \mid \bm\beta_{k^a}^{a}}))
\\
\mathbf{KL}(\bm\beta_{k^a}^{a}\mid \mid \bm\beta_{k^b}^{b}) =\frac{1}{N_{v}}\sum_{v=1}^{N_{v}}\bm\beta_{k^a v}^{a}\cdot\log\left(\frac{\bm\beta_{k^a v}^{a}}{\bm\beta_{k^b v}^{b}} \right) \\
\end{multlined}$}
\end{equation}

\noindent Given a threshold $\tau$ the similarity between two
topics can be binarized. Topic pairs with distance less than a
threshold are counted as inliers, defined by:

\begin{equation}\label{eq:Inlier}
\resizebox{.8\hsize}{!}
{$
\begin{multlined}
NumInlier = \sum\limits_{k^{a}}\mathbbm{1}(\min\limits_{k^b}(\mathcal{D}_{\mathbf{KL}}(\bm\beta_{k^{a}}^{{a}},\bm\beta_{k^{b}}^{{b}}))<\tau)\\+\sum\limits_{k^{b}}\mathbbm{1}(\min\limits_{k^a}(\mathcal{D}_{\mathbf{KL}}(\bm\beta_{k^{b}}^{{b}},\bm\beta_{k^{a}}^{{a}}))<\tau)
\end{multlined}$}
\end{equation}

\noindent where $\mathbbm{1}(\cdot)$ is the indicator function. The final relatedness measure $\mathcal{D}({a},{b})$ between scenes $a$ and $b$ is the percentage of inlier topic pairs:

\begin{equation}\label{eq:SceneSimilarity}
\mathcal{D}({a},{b}) = \frac{NumInlier}{K^a+K^b}
\end{equation}

Since Eqs.~\ref{eq:SymKLDivergence} and \ref{eq:Inlier} are symmetric, Eq.~\ref{eq:SceneSimilarity} is as well. Given this relatedness measure, every scene pair is compared to generate an affinity matrix, and self-tuning spectral clustering \cite{conf/nips/Zelnik-ManorP04} is used to group scenes into $c=1\dots C$ semantically similar scene-level clusters. (See Fig.~\ref{Fig:MultilayerClustering}~II for an example).

\subsection{Learning A Shared Activity Topic Basis}\label{subsection:LearningSTB}

Scenes clustered according to Section~\ref{subsection:SceneLevelClustering} are semantically similar, however the representation in each is still distinct. We next show how to establish a shared representation for every scene in a particular cluster. We denote the set of scenes in a cluster as $\mathcal{C}$.
We first choose the scene with the lowest distance to all other scenes in the cluster as the reference scene/coordinate $s_{ref}$. Activities in all scenes $s\in\mathcal{C}$ can be projected to the reference coordinates via transform $\mathbf{T}^{s2s_{ref}}$ as stated in Eq.~(\ref{eq:TopicProject}).
\begin{equation}\label{eq:TopicProject}
\forall s\in \mathcal{C}, \forall k=1\dots K : \tilde{\bm\beta}_{k}^{s}=\mathbb{H}(\bm\beta_{k}^{s};\mathbf{T}^{s2s_{ref}})
\end{equation}

Once every topic is in the same coordinate system, we create an affinity matrix for all the transformed topics $\{\bm\tilde{\bm\beta}^{s}_k\}_{s\in \mathcal{C}}$
using the symmetrical Kullbeck-Leibler Divergence as distance metric (Eq.~(\ref{eq:SymKLDivergence})).
Hierarchical clustering is then applied to group the projected activities into $K^{stb}$ clusters $\{\mathcal{T}_k\}_{k=1}^{K^{stb}}$. ($\mathcal{T}_k$ denotes the set of activities in a cluster $k$). The result is that semantically corresponding activities across scenes are now grouped into the same cluster.
We then take the mean of activities in each activity cluster $\mathcal{T}_k$ as one {\em shared activity topic} $\bm\beta_{k}^{stb}$ as in
Eq.~(\ref{eq:STB}). {An alternative to this approach is
  to re-learn topics from the concatenation of visual words of all
  the scenes in a single cluster. However, this `Learning-from-Scratch'
  strategy prevents explicitly identifying shared and unique topics
  across scenes. Because the trace of local topics from individual
  scenes to STB is lost. In contrast, our framework reveals how scenes
  are similar or different.} 

\begin{equation}\label{eq:STB}
\forall k=1\dots K^{stb} : \bm\beta_{k}^{stb}=\frac{1}{\mid \mathcal{T}_k \mid}\sum\limits_{k',s'\in \mathcal{T}_k} \bm\tilde{\bm\beta}_{k'}^{s'}
\end{equation}

We denote the set of {\em shared activity topics} $\{\bm\beta_{k}^{stb}\}_{k=1}^{K^{stb}}$ learned for the cluster as the {\em shared activity topic basis} (STB). The resulting STB captures both common and unique activities in every scene member. See Fig.~\ref{Fig:MultilayerClustering}III for an example. We can now represent the behaviours in every scene as STB profiles: by projecting the STB back to each scene and re-computing the topic profile $\bm\gamma^{stb}_{j}$ defined now on $\{\bm\beta_{k}^{stb}\}_{k=1}^{K^{stb}}$; in contrast to the original scene-specific representation ($\bm\gamma_j^s$, defined in terms of $\{\bm\beta^{s}_k\}_{k=1}^{K}$). 
{That is, re-running Algorithm \ref{Alg:ModelInference}, but with $\bm\beta$ fixed to the STB values obtained from Eq.~(\ref{eq:STB}).}  An example of behaviour profiling on STB is illustrated in Fig. \ref{fig:BehaviourProfiling}. Visual words accumulated within a clip are profiled according to the STB. Thus each behaviour can be treated as a weighted mixture of multiple activities.

\begin{figure*}[!htb]
\begin{center}

\includegraphics[width=0.8\linewidth]{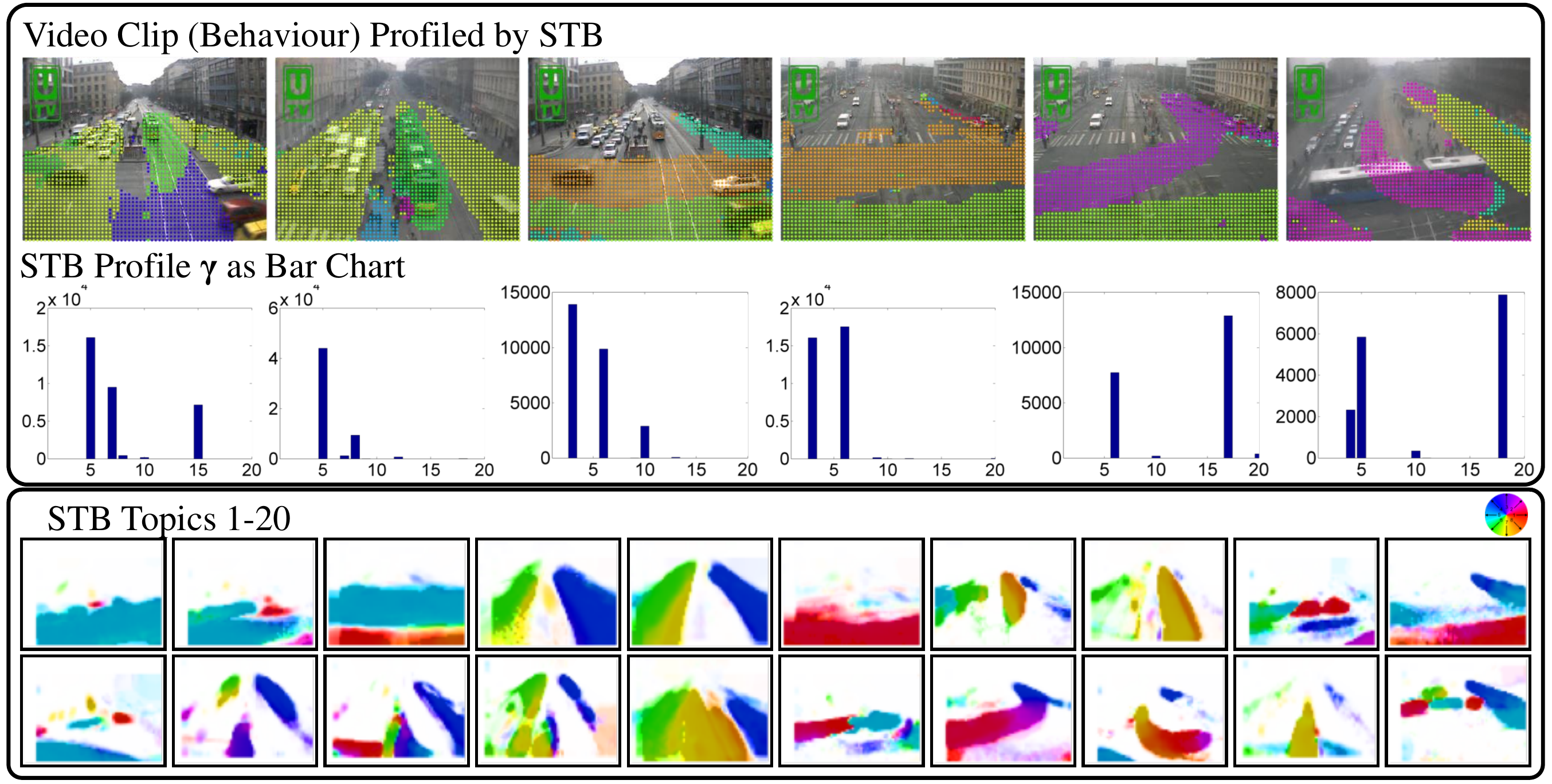}
\caption{An illustration of behaviour profiling on STB. In the left block, visual words are profiled by STB and plotted as coloured dots. Notice that colors here indicate visual words belonging to individual activities in STB instead of motion direction. Profiling $\gamma$ is also given as bar chart where x axis indexes STB activities. The right block illustrates the STB activities where color patches indicate distribution of motion vectors.}
\label{fig:BehaviourProfiling}
\end{center}
\vspace{-0.5cm}
\end{figure*} 

\section{Cross-Scene Query by Example and Classification} \label{section:Query&Classification}
Given the structured multi-scene model introduced in the previous section, we can now describe how cross-scene query and classification can be achieved.\\

\noindent\textbf{Cross-scene query}\quad Activity-based query by example aims at retrieving semantically similar clips to a given query clip. In the cross-scene context, the pool of potential clips to be searched for retrieval includes clips from every camera in the network.
Within a scene cluster $\mathcal{C}$, we segment each video
$s$ into $j=1\dots N_{d}$ short clips
(Section~\ref{subsection:VideoClipRepresentation}). We represent the
$jth$ video clip in scene $s$ as topic profile $\bm\gamma^{stb}_{js}$
defined on STB $\bm\beta_k^{stb}$. A query clip $q$, represented by STB
profile $\bm\gamma^{stb}_{qs}$ can now be directly compared against all
other clips in the cluster $\{\bm\gamma^{stb}_{js'}\}_{j,s'\in \mathcal{C}}$ using L2 distance. In this way, \emph{cross-scene query-by-example} is achieved by sorting all clips in the cluster according to distance to the query. \\

\noindent\textbf{Cross-scene classification}\quad Given an existing annotated database of scenes modelled with our multi-layer framework, classification in a new scene $s^*$ can now be achieved \emph{without further annotation}. First $s^*$ is associated to a cluster $c^*$ (Section~\ref{subsection:SceneLevelClustering}). Although $s^*$ has no annotation, this reveals a set of semantically corresponding existing scenes from which annotation can meaningfully be borrowed. Classification can thus be achieved by any classifier, using all other scenes/clips and labels from cluster $c^*$ as the labeled training set.

{It should be noted that our cross-scene classification differs
  from \cite{DBLP:conf/iccv/ZhengJ13,ZhengJPC_BMVC12} in: (1) We train on a
  \textbf{set} of source scenes before testing on a held-out scene
  rather than one source to one test scene. The conventional 1-1
  approach requires implicitly the source and target scene to be {\em
    relevant} which must be manually identified. Our model is able to
  group relevant 
  scenes automatically without requiring the user to know this as {\em
    a priori}. (2) Our model works in a transductive
  \cite{journal/IKDE/PanY2010} manner. That is, it looks at target
  scene data during scene clustering, but without looking at the
  target data label. This weak assumption is more desirable in
  practice because surveillance video data is often easy to collect
  but without any labelling, whilst the effort required for labelling
  is the bottleneck. }

\section{Multi-Scene Summarization} \label{Subsection:MultiVideoSummarization}
In this section we present a multi-scene video summarization algorithm that exploits  the structure learned in Section~\ref{sec:sceneActCluster} to compress cross-scene redundancy. All clips are represented by their profile on
STB. The general objective of multi-scene summarization 
is to generate a \textit{video skim} with at least one example of each
distinct behaviour in the shortest possible summary. We generate
independent summaries for each  scene cluster (since different scene clusters are
semantically dissimilar), and multi-scene summaries within each
cluster (since scenes within a cluster are semantically similar). \\ 

\noindent\textbf{K-center summaries:}
The multi-scene summary video is of configurable length $N_{sum}$. Longer videos will show more distinct behaviours or more within-class variability of each behaviour.  We compose the summary $\Sigma$ of  $N_{sum}$ clips $\{\bm\gamma^{stb}_{j}\},{j\in \Sigma}$ drawn from all scenes in the cluster.  The objective is that all clips in the cluster $\{\bm\gamma_{js}^{stb}\}_{j,s\in\mathcal{C}}$ should be near to at least one clip in the summary (i.e., the summary is representative). Formally, this objective is to find the summary set $\Sigma$ that minimizes the cost $J$  in Eq.~(\ref{eq:KcenterObjective}) {where $\mathcal{D}_{\gamma}$ is the L2 distance}: 
\vspace{-0.1cm}
\begin{equation} \label{eq:KcenterObjective}
J = \underset{j,s\in\mathcal{C}}{max} \left( \underset{j'\in\Sigma}{max} \quad \mathcal{D}_{\gamma}\left(\bm\gamma^{stb}_{j'},\bm\gamma_{js}^{stb}\right) \right)
\end{equation}

This is essentially a k-center problem
\cite{journals/tcs/Gonzalez85}. Since it is intractable  to enumerate 
all combinations/potential summaries $\Sigma$, we adopt the
2-approximation algorithm \cite{Hochbaum1985} to this
optimization. The resulting $K=N_{sum}$ centers identify the summary clips. 

\section{Experiments}
\noindent\textbf{Dataset:}\quad We collected 25 real traffic surveillance videos from publicly accessible online web-cameras  in Budapest, Hungary. 
These videos are combined with two surveillance video datasets
 Junction and Roundabout~\cite{DBLP:journals/ijcv/LiGX12} for a
total of 27 videos. Sample frames for each scene are illustrated in Fig. \ref{Fig:SampleFrames}(a). We trim each video to 18 000 frames in
10fps, of which 9 000 are used to learn the model and the remaining 9 000 frames are used for testing (query, classification and summarization). For activity learning we segment each training video into 25 frame clips, so 360 clips are generated for each scene. For both query and summarization
  applications, we segment test videos into clips with 80 frames, so
112 clips for query and summarization are generated from each
scene. {Thus, we have three types of video clips: (1) Clips for
  unsupervised training of LDA, (2) clips for training cross-scene
  classification, retrieval and multi-scene summarization, (Semantic
  Training Clips), (3) clips for testing cross-the same tasks
  (Semantic Testing Clips). LDA clips are shorter (25 frames) to
  facilitate learning more cleanly segmented activities. Semantic
  clips are longer (80 frames) as a more human-scale user-friendly unit for
  visualisation and annotation. } 

\begin{figure*}[!ht]
\begin{center}
{\includegraphics[width = 0.95\textwidth]{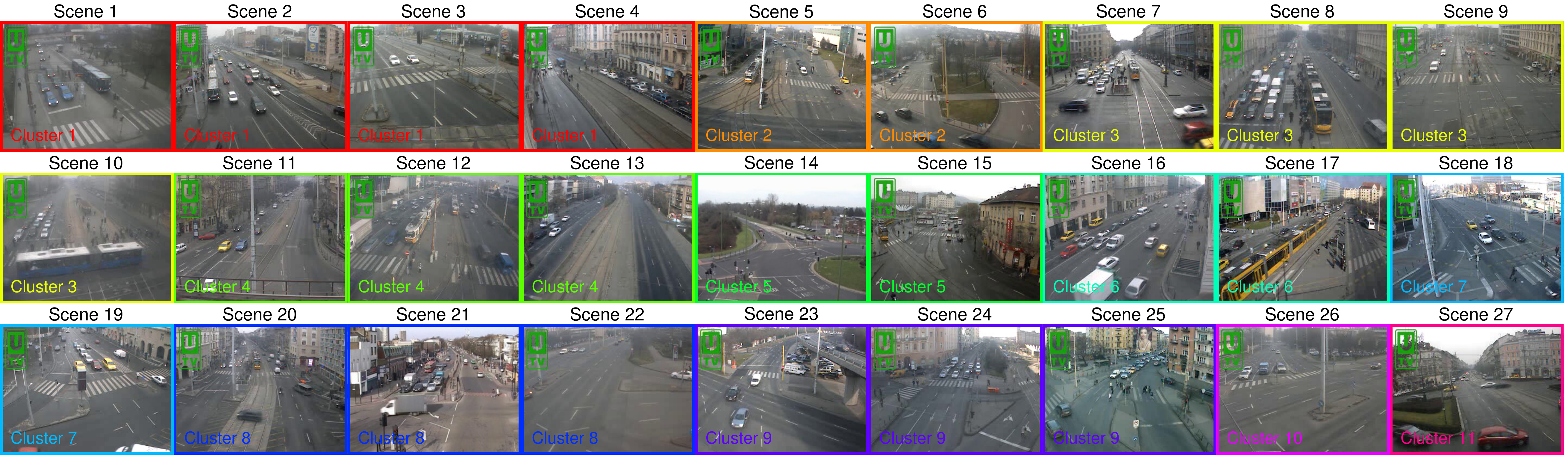}}
\end{center}
\caption{Example frames for our multi-surveillance video dataset with each scene assigned a reference number on top of the frame. The color of bounding box and text in the bottom left indicates assigned cluster.}
\label{Fig:SampleFrames}
\vspace{-0.6cm}
\end{figure*}

\noindent\textbf{Learning Activities:}\quad We computed optical flow
\cite{CLiu:BP} for all videos by quantizing the scenes
with $5\times 5$ pixel cells and 8 directions. Local
activities are learned from each video independently using LDA with $K=15$ activities per scene.\\ 
\noindent\textbf{Behaviour Annotation:}\quad Behaviour is a clip-level semantic tag defining the overall scene-activity. Due to the semantic gap between behaviours in the video clip and (potentially
task dependent) human interpretation, it is difficult to give video a concise and consistent semantic label (in contrast to human action  \cite{DBLP:conf/iccv/ZhengJ13} and event \cite{conf/iccv/KhokharSS11} recognition). Instead of annotating each video clip explicitly, we give a
set of binary activity tags (each representing the action of some
objects within the scene) to each video clip as shown in
Table~\ref{Tab:AnnotationScheme}. \added{All the tags associated with
  vehicles have a sparse or dense option. 
When there are less than three vehicles travelling in a clip, it is labelled as sparse,
otherwise dense.} Each unique
combination of activities that exists in the labelled clips  then defines a unique
scene-level behaviour category. 
We explore this
through multiple sets of annotations: an original annotation with 19 distinct tags, and subsequent coarser  label sets derived by merge scheme 1 with 13 distinct tags and merge scheme 2 with 10 distinct tags. The activity tags are given in Table~\ref{Tab:AnnotationScheme}. We exhaustively annotate video clips in two example scene clusters (3 and
7 as shown in Fig. \ref{Fig:SampleFrames}). {Across the two clusters, there are 6
scenes with 112 clips per scene annotated (672 clips in total).} In the original annotation case, there are 111
total behaviours identified. The distribution of behaviours are illustrated in Fig.~\ref{fig:BehavioursFrequency}(a). However this number is more than necessary in terms
of limited distinctiveness of the numerous entailed behaviours. By merging some
activity annotations we generate 59 or 31 (Merge
Scheme 1 or 2 in Table~ \ref{Tab:AnnotationScheme}) unique
behaviours. It should be noted that the frequency of 
behaviours is rather imbalanced, as indicated by all the subfigures
of Fig.~\ref{fig:BehavioursFrequency}. 
There is also very limited overlap of behaviours between scene clusters
3 and 7. \added{To assess annotation consistency and bias, we invited
  eight independent annotators to annotate all the video clips
  separately. We observe that the additional annotations are fairly 
  consistent with the original annotation: with more than $80\%$ agreement
  (Hamming distance) between the additional and the original
  annotations. Detailed analysis of these additional annotations are
  given in the supplementary material.}

\begin{figure}
\begin{center}
\subfloat[Behaviour frequency: original annotation]{\includegraphics[width=0.95\linewidth]{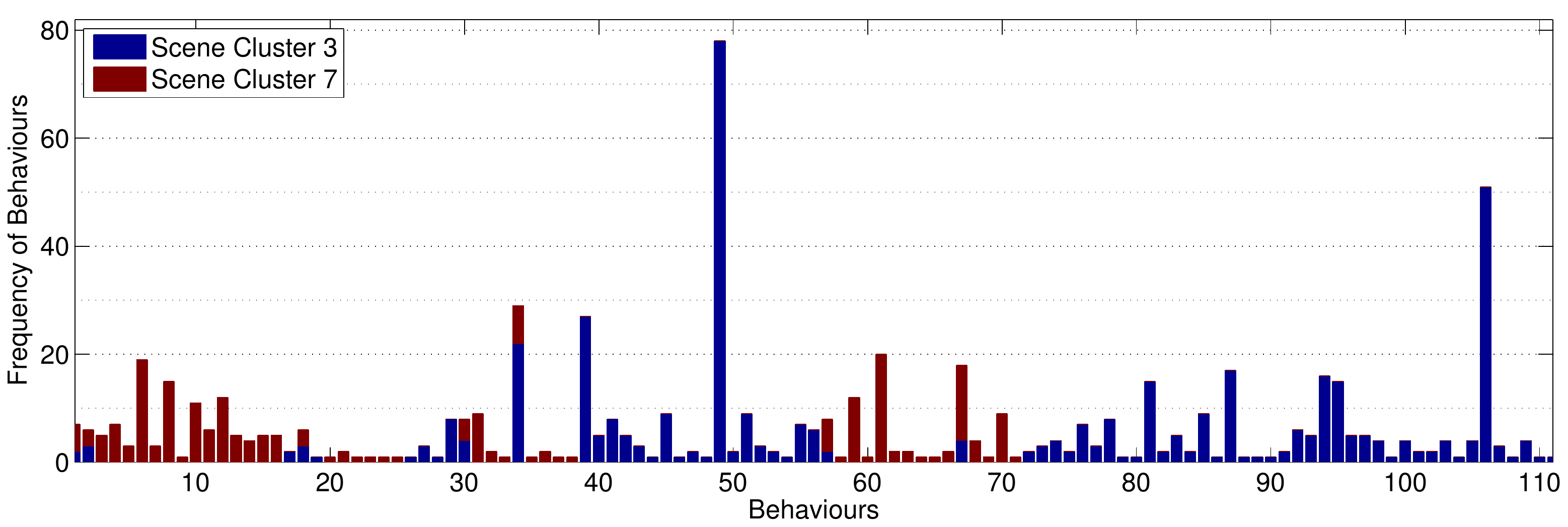}}\\
\subfloat[Behaviour frequency: merge scheme 1]{\includegraphics[width=0.61\linewidth]{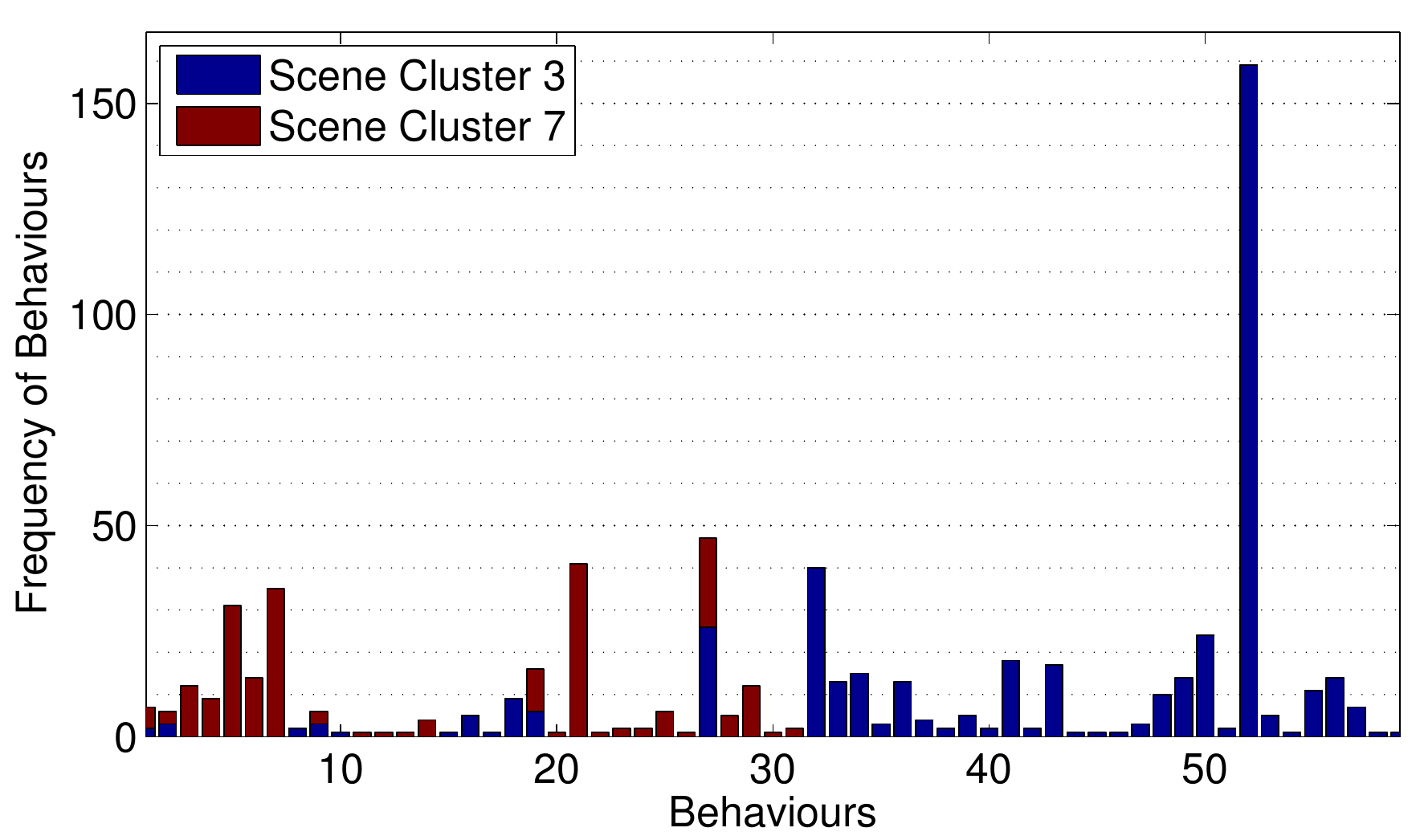}}
\hspace{0.1cm}
\subfloat[Behaviour frequency: merge scheme 2]{\includegraphics[width=0.34\linewidth]{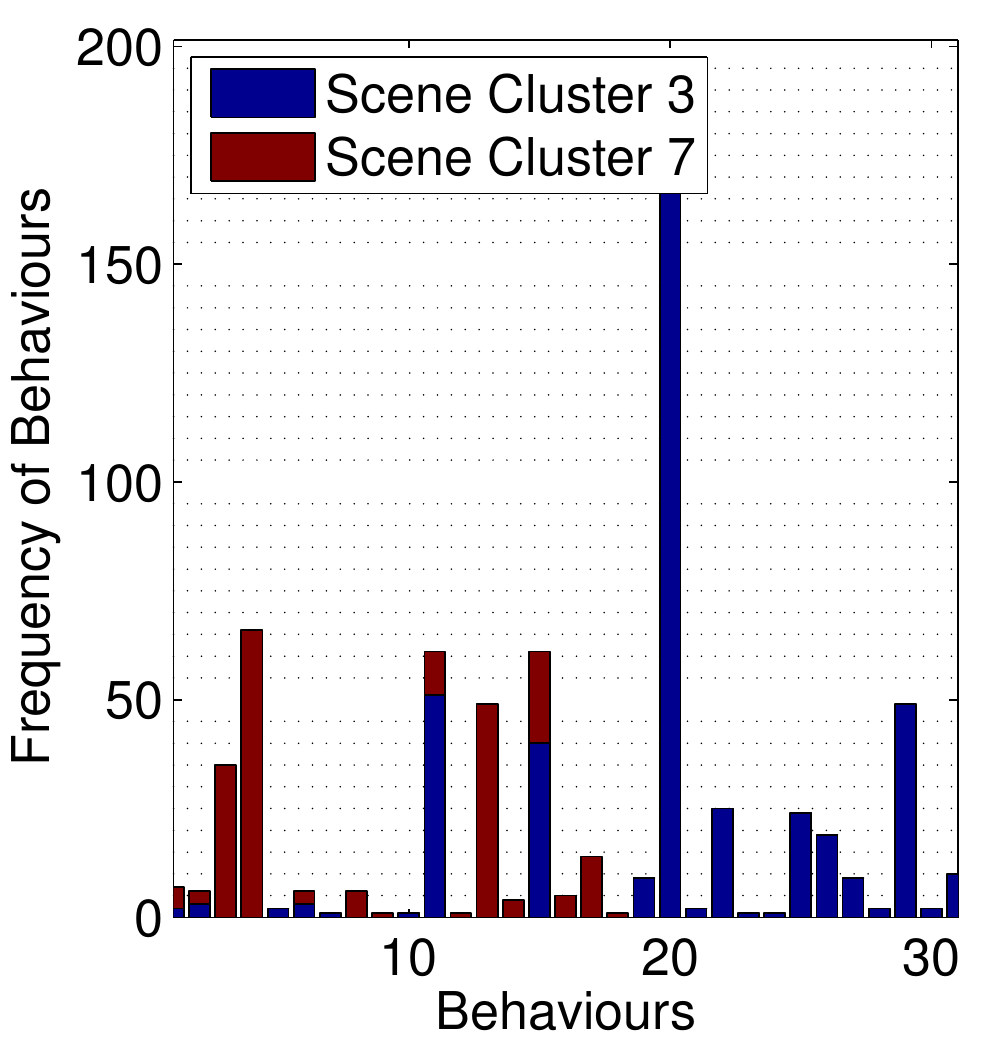}}
\end{center}
\caption{Frequencies of behaviours of each category. (a), (b) and (c) illustrate the frequency of behaviours when varying the labelling criteria. }\label{fig:BehavioursFrequency}
\end{figure}

\begin{table}[htb]
\centering
\caption{Original annotation ontology and two merging schemes give multiple granularities of annotation.}
\label{Tab:AnnotationScheme}
\resizebox{0.49\textwidth}{!}{
\begin{tabular}{p{0.02cm}|p{2.9cm}|p{2.7cm}|p{2.7cm}}
\toprule
\multicolumn{1}{l|}{No.} & \multicolumn{1}{c|}{Original Annotation} & \multicolumn{1}{c|}{Merge Scheme 1} & \multicolumn{1}{c}{Merge Scheme 2} \\ \midrule
1 & Vehicle Left Sparse & \multicolumn{ 1}{l|}{Vehicle Left} & \multicolumn{ 1}{l}{Vehicle Horizontal} \\ \cline{ 1- 2}
2 & Vehicle Left Dense & \multicolumn{ 1}{l|}{} & \multicolumn{ 1}{l}{} \\ \cline{ 1- 3}
3 & Vehicle Right Sparse & \multicolumn{ 1}{l|}{Vehicle Right} & \multicolumn{ 1}{l}{} \\ \cline{ 1- 2}
4 & Vehicle Right Dense & \multicolumn{ 1}{l|}{} & \multicolumn{ 1}{l}{} \\ \hline
5 & Vehicle Up Sparse & \multicolumn{ 1}{l|}{Vehicle Up} & \multicolumn{ 1}{l}{Vehicle Vertical} \\ \cline{ 1- 2}
6 & Vehicle Up Dense & \multicolumn{ 1}{l|}{} & \multicolumn{ 1}{l}{} \\ \cline{ 1- 3}
7 & Vehicle Down Sparse & \multicolumn{ 1}{l|}{Vehicle Down} & \multicolumn{ 1}{l}{} \\ \cline{ 1- 2}
8 & Vehicle Down Dense & \multicolumn{ 1}{l|}{} & \multicolumn{ 1}{l}{} \\ \hline
9 & Vehicle Southeast Sparse & \multicolumn{ 1}{l|}{Vehicle Southeast} & \multicolumn{ 1}{l}{Vehicle SE\& NW} \\ \cline{ 1- 2}
10 & Vehicle Southeast Dense & \multicolumn{ 1}{l|}{} & \multicolumn{ 1}{l}{} \\ \cline{ 1- 3}
11 & Vehicle Northwest Sparse & \multicolumn{ 1}{l|}{Vehicle Northwest} & \multicolumn{ 1}{l}{} \\ \cline{ 1- 2}
12 & Vehicle Northwest Dense & \multicolumn{ 1}{l|}{} & \multicolumn{ 1}{l}{} \\ \hline
13 & Vehicle Up2Right Turn & Vehicle Up2Right Turn & Vehicle Up2Right Turn \\ \hline
14 & Vehicle Left2Up Turn & Vehicle Left2Up Turn & Vehicle Left2Up Turn \\ \hline
15 & Vehicle Up2Left Turn & Vehicle Up2Left Turn & Vehicle Up2Left Turn \\ \hline
16 & Tram Up & Tram Up & Tram Up \\ \hline
17 & Tram Down & Tram Down & Tram Down \\ \hline
18 & Pedestrian Horizontal & Pedestrian Horizontal & Pedestrian Horizontal \\ \hline
19 & Pedestrian Vertical & Pedestrian Vertical & Pedestrian Vertical\\ \bottomrule
\end{tabular}}
\vspace{-0.5cm}
\end{table}

\subsection{Multi-Layer Scene Clustering}
\noindent\textbf{Scene Level Clustering:} \label{subsection:SceneLevelClusteringResult}
We first group the scenes into semantically similar clusters by spectral clustering. The similarity measurement between scenes is the number of corresponding activities, as defined in Section~\ref{subsection:SceneLevelClustering}. The self-tuning spectral clustering automatically determines the appropriate number of clusters which, in the case of our 27-scene dataset, is 11 clusters. Fig.~\ref{Fig:SampleFrames} shows the results, in which semantically similar scenes are
indeed grouped (e.g. Camera towards one direction at road junctions in Cluster 3), and unique views
are separated into their own cluster (e.g. Cluster 11).\\

\noindent\textbf{Learning A Shared Activity Topic Representation:}\quad 
Within each scene cluster we unify the representation by computing a {\em shared activity topic
  basis}. We automatically set the number of shared activities
$K^{stb}$ in each scene cluster with $N_{s}$ scenes as
$K^{stb}=coeff\times N_{s}$ where $coeff$ is set to 5. The discovered basis from an
example cluster {(Scene Cluster 3 shown in Fig.\ref{Fig:SampleFrames})} with 4 scene members is illustrated in Fig.~\ref{fig:TopicCluster}. This figure reveals both activities unique to each scene (Topics 1-15) and activities common among multiple scenes (Topic 16-20). Thus some shared activity topics are composed of single local/original topics, and others of multiple local topics.

\begin{figure*}[!htb]
\begin{center}
	\includegraphics[width=0.99\linewidth]{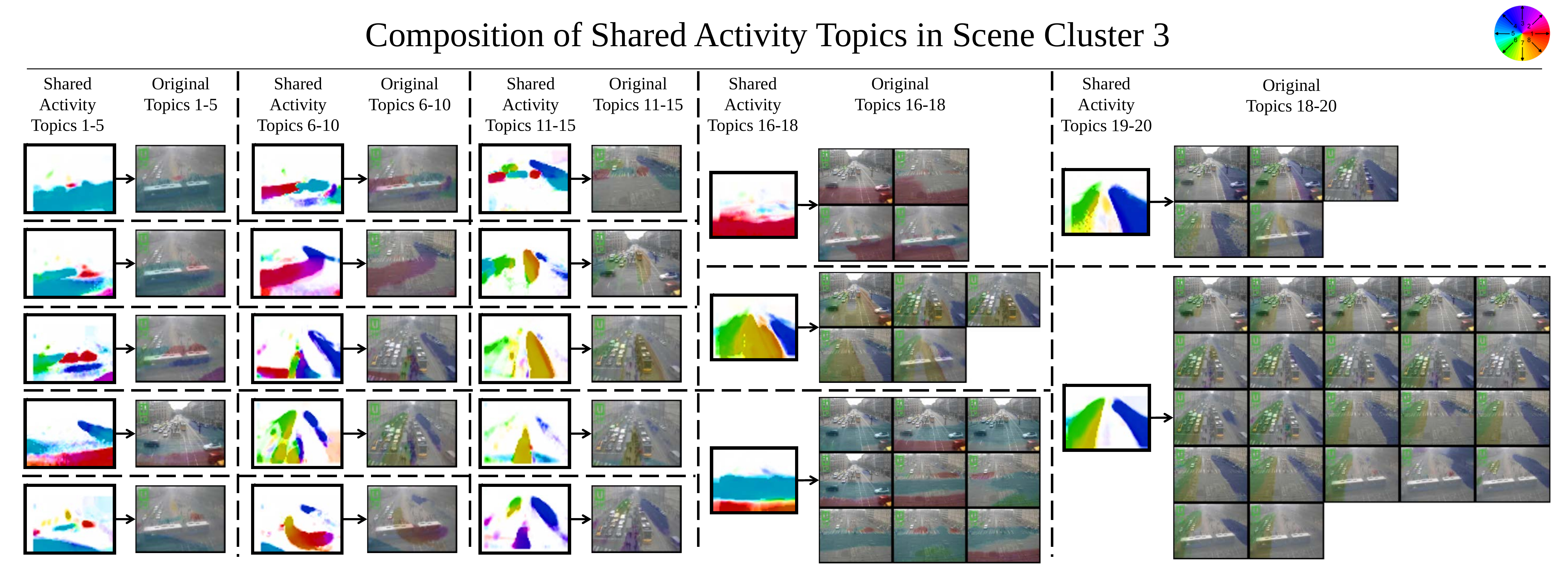}
\caption{Example STB learned from Scene Cluster 3. Shared activity topics may be composed of one or more local/original topics. Original topics are overlaid on background frame. Color patches indicate distribution of motion vectors for a single activity.}\label{fig:TopicCluster} 
\end{center}
\vspace{-0.5cm}
\end{figure*}

\subsection{Cross-Scene Query by Example and Classification}\label{sect:Exp_QBE_Classification}

In this section we evaluate the ability of our framework to support
two tasks: cross-scene query by example; and cross-scene behaviour classification. We compare our
\textbf{Scene Cluster Model} (SCM) with a baseline \textbf{Flat Model} (FM). Our \textbf{Scene Cluster Model} first group scenes into scene clusters according to their relatedness and learns STB for every scene cluster. Video clips in each scene cluster are thus represented as topic profiles on the STB of the scene cluster. As with our model, a \textbf{Flat Model} first learns a local topic model per scene,
however it then learns a single STB from all labelled scenes (6 scenes from 2 clusters) without scene level clustering,
instead of one STB per-cluster. {The only difference between SCM and FM is the absence of scene-level clustering in FM. Note that the Flat Model is a special
case of our Scene Cluster Model with 1 scene-level cluster. Moreover, 
the individual scenes are also a special case of our Scene Cluster
Model with one cluster per scene.}\\

\noindent\textbf{Query by Example Evaluation:}  To quantitatively evaluate query by example, we exhaustively take each scene and each clip in turn as the query, {and all other scenes are considered as the pool}. All clips in the pool are ranked according to similarity (L2 distance on STB profile) to the query. Performance is evaluated according to how many clips with the same behaviour as the query clip are in the top $T$ responses.
{We retrieve the best $T = 1\cdots 200$ clips and calculate the
  {\em Average Precision} of each category for each $T$. MAP is
  computed by taking the mean value of {\em Average Precision} over
  all categories. The MAP curve by the top T responses to a query for
  both \textbf{Scene Cluster Model} (SCM) and \textbf{Flat Model} (FM)
  and Merge Scheme 1 and 2 are plotted in Fig.~\ref{fig:QBE_T}. 
\begin{figure}
\begin{center}
\includegraphics[width = 0.80\linewidth]{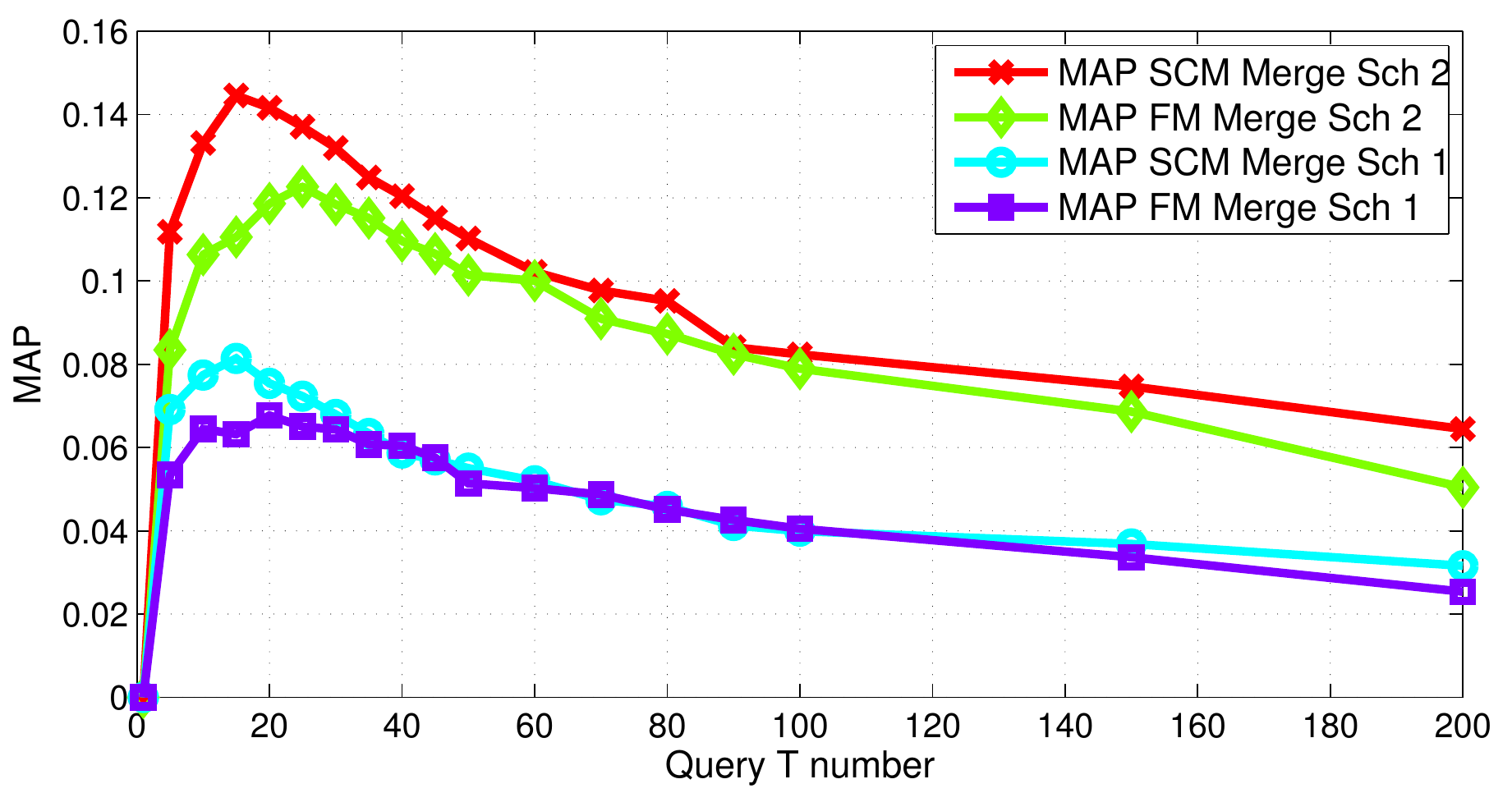}
\caption{Query by example MAP with different number of retrievals}\label{fig:QBE_T}
\end{center}
\vspace{-0.6cm}
\end{figure}
It is evident that for both Merge Scheme 1 and 2, the proposed scene cluster
model (SCM) performs consistently better than the Flat Model (FM)
regardless of number of top retrievals T.
This is because in the Scene Cluster Model, the STB learned from this
set of scenes are highly relevant to each scene in the cluster. In
contrast, the Flat Model learns a single STB for all scenes making the STB
less relevant to each individual scene, hence less informative as
a representation for retrieval. }

Qualitative results are also given in Fig.~\ref{fig:QBE_Qualitative} by presenting 6 randomly chosen queries and their retrieved clips. Different types of behaviours are covered by query clips and most retrieved clips are semantically similar to query clips. The only exception is in the 3rd row where the query clip indicates traffic going east and turning from left to up. This is because there is no corresponding behaviour in the other scenes.\\

\begin{figure}[!t]
\begin{center}
\includegraphics[width=0.95\linewidth]{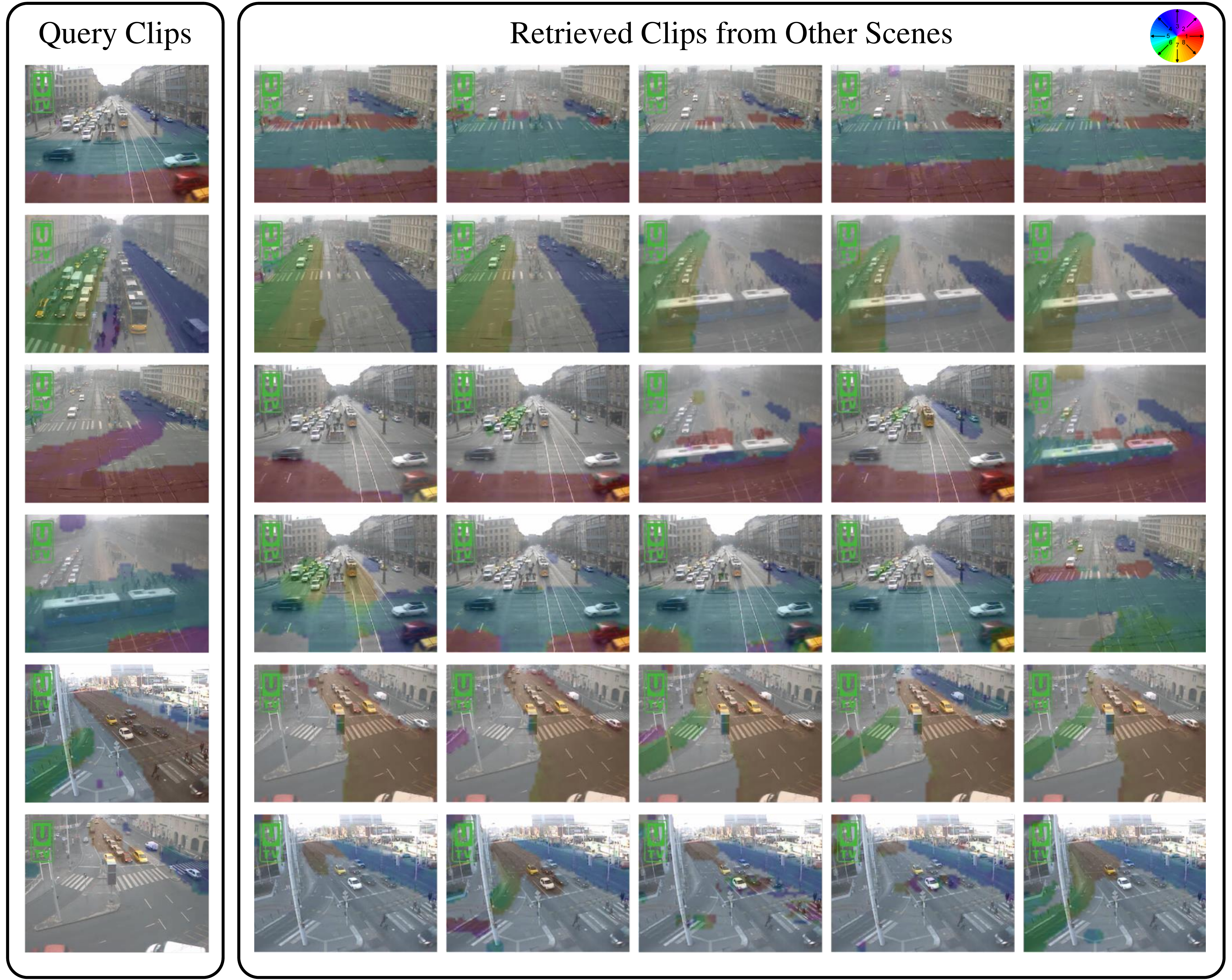}
\end{center}\caption{Examples of cross-scene query by example. The first column
  gives 6 query clips randomly chosen from 6 scenes. The right image
  matrix illustrates the retrieved clips from the remaining 5 scenes, sorted by distance to
  query from left to right in the matrix. Color patches overlaid on the 
  background indicates the visual words accumulated within a video
  clip.} \label{fig:QBE_Qualitative}
\end{figure}

\noindent\textbf{Classification Evaluation:} In this experiment we quantitatively evaluate classification performance where the test scene has \emph{no labels}. Successful classification thus depends correctly finding semantically related scenes and appropriately transferring labels from them (Section~\ref{section:Query&Classification}). We perform leave one scene out evaluation by holding out one scene as the unlabelled testing set, and predicting the labels for the test set clips using the labels in remaining scenes using the KNN classifier. The KNN K parameter is determined by cross validating among the remaining scenes.
Classification performance is evaluated by the accuracy for each category of behaviour, averaged over all held out scenes.

\begin{table}[!hb]
\centering
\caption{Cross-scene classification accuracy with 31 and 59 categories for both Scene Cluster Model (SCM) and Flat Model (FM).}
\resizebox{0.45\textwidth}{!}{
\begin{tabular}{p{1.1cm}|p{1.4cm}|p{1.4cm}||p{1.4cm}|p{1.4cm}}
\toprule
Category & \multicolumn{2}{c||}{31}              & \multicolumn{2}{c}{59}              \\ \cline{2-5} 
         & SCM & FM       & SCM & FM       \\ 
\midrule
Scene 1  & \textbf{55.36\%}  & 50.89\%          & \textbf{42.86\%}  & 40.18\%          \\
Scene 2  & 27.68\%           & \textbf{39.29\%} & \textbf{18.75\%}  & 16.96\%          \\
Scene 3  & \textbf{49.11\%}  & 41.96\%          & \textbf{39.29\%}  & 37.50\%          \\
Scene 4  & \textbf{54.46\%}  & 46.43\%          & \textbf{37.50\%}  & 36.61\%          \\
Scene 5  & \textbf{30.36\%}  & 26.79\%          & \textbf{17.86\%}  & \textbf{17.86\%} \\
Scene 6  & \textbf{38.39\%}  & 25.00\%          & \textbf{20.54\%}  & 12.50\%          \\
Average  & \textbf{42.56\%}  & 38.39\%          & \textbf{29.47\%}  & 26.94\%          \\ \hline
\end{tabular}}
  \label{Tab:ClassificationAccuracy}
\end{table}

From Table~\ref{Tab:ClassificationAccuracy} we observe that at either
granularity of annotation (59 or 31 categories), our \textbf{Scene
Cluster Model} outperforms the \textbf{Flat Model} on average. This shows that
again in order to  borrow labels from other scenes for cross-scene
classification, it is important to select relevant sources, which we achieve via scene
clustering. The \textbf{Flat Model} is easily confused by the
wider variety of scenes to borrow labels from, while our \textbf{Scene 
Cluster Model} structures similar scenes and borrows labels from only semantic related scenes to avoid `negative transfer'
\cite{journal/IKDE/PanY2010,Xu:2013:CTS:2510650.2510657}.

\subsection{Multi-Scene Summarization}\label{sect:Exp_MultiSceneSummary}
In the final experiment, we evaluate our multi-scene summarization model against a variety of alternatives. We consider two conditions: In the first, we consider multi-scene summarization within a scene cluster (Condition WC); in the second we consider unconstrained multi-scene summarization including videos spanning multiple scene clusters (Condition AC).

\vspace{0.1cm}\noindent{\textbf{Condition: Within-cluster summarization (WC)}}\quad In this experiment we focus on the comparison between \textbf{Multi-Scene Model} and \textbf{Single-Scene Model} given various summarization algorithms. The \textbf{Multi-Scene Model} represents all video clips from different scenes within a cluster with a single STB learned from the scene cluster while the \textbf{Single-Scene Model} represents each video with scene specific activities and the overall summary is the mere concatenation of summaries from each scene. Specifically, we compare the summarization methods listed in Table~\ref{tab:SummSchWC}.

\begin{table}[!h]
  \centering
  \caption{Summarization schemes for \textbf{Condition WC}}\label{tab:SummSchWC}
  \resizebox{0.42\textwidth}{!}{
    \begin{tabular}{p{2cm}|p{6cm}}
    \toprule
    \textbf{Summarization Method} & \multicolumn{1}{c}{\textbf{Description}} \\
    \midrule
    \textit{Random} & {This lower-bound picks clips randomly from multiple scenes to compose the summary} \\
    \hline
    \textit{Single-Scene Graph} & {The overall summary is a concatenation of independent summaries for each video by doing recursive Normalized cut \cite{Shi00normalizedCuts} on a graph constructed by taking each video clip as vertices and L2 distance between topic profile $\gamma$ of each clip as edges. Here each video clip is represented by scene-specific local topics. {This corresponds to \cite{journals/tcsv/NgoMZ05}, but without temporal graph.}}  \\
        \hline
    \textit{Single-Scene Kcenter} & {Similar to \textit{Single-Scene Graph} method, but using Kcenter algorithm in Eq.~(\ref{eq:KcenterObjective}) for summarization instead of Normalized Cut.} \\
        \hline
    \textit{Multi-Scene Graph} & {This model learns a STB to represent video clips from all scenes with STB profile. Then Normalized Cut is applied to cluster clips and find multi-scene summaries.} \\
        \hline
    \textit{Multi-Scene Kcenter} & {Our full model builds a STB from all scenes within a cluster, then uses the Kcenter algorithm to select summary clips from all scenes.} \\
    \bottomrule

    \end{tabular}
  }
\end{table}

\vspace{0.1cm}\noindent{\textbf{Condition: Across-cluster summarization (AC)}}\quad In this experiment, analogous to query and classification, we focus on the comparison between \textbf{Flat Model} and \textbf{Scene Cluster Model} given different summarization algorithms. The \textbf{Flat Model} Learns a single STB from all scenes available without discrimination while \textbf{Scene Cluster Model} learns a STB per scene cluster. 
Specifically, we compare the summarization schemes in Table~\ref{tab:SummSchAC}.

\begin{table}[!h]
  \centering
  \caption{Summarization schemes for \textbf{Condition AC}}\label{tab:SummSchAC}
    \resizebox{0.42\textwidth}{!}{
    \begin{tabular}{p{2cm}|p{6cm}}
    \toprule
    \textbf{Summarization Method} & \multicolumn{1}{c}{\textbf{Description}} \\
    \midrule
    \textit{Random} & {This picks clips randomly from multiple scenes to compose the summary} \\
    \hline
    
    \textit{Flat Multi-Scene User Attention} & {{Leverages the magnitude, spatial and temporal phase of optical flow vectors to index videos. This is the visual attention measurement  of (\cite{journals/tmm/MaHLZ05}, Eq.~(6)). We tested the model on a combined video by concatenating each individual video.}} \\
\hline    
    
    \textit{Flat Multi-Scene Graph} & {{This model uses Normalized Cut \cite{Shi00normalizedCuts} to cluster all video clips represented as single STB profiles. This is similar to \cite{journals/tcsv/NgoMZ05}.}} \\
    \hline
    
    \textit{Flat Multi-Scene Kcenter} & {Same as \textit{Flat Multi-Scene Graph}, but using Kcenter to select summary clips.} \\
    \hline
    
    \textit{Scene Cluster Multi-Scene Kcenter} & {Our full model clusters the scenes, learns STBs on each scene cluster, followed by Kcenter to summaries within each scene cluster} \\
    \bottomrule
    \end{tabular}
  }
\end{table}

\vspace{0.2cm}\noindent\textbf{Settings:}\quad 
To systematically evaluate summarization performance, we vary the length of the requested summary. In \textbf{Condition WC} the  summary varies from 8 to 120 clips (64seconds to 16mins) out of overall 448 video clips (59.7mins) in Scene Cluster 3 (as shown in Fig.~\ref{Fig:SampleFrames}(a)) and 224 video clips (29.9mins) in Scene Cluster 7. In \textbf{Condition AC} the summary varies from 6 to 120 clips (48seconds to 16mins) out of 672 video clips (89.7mins) total  which is a combination of Scene Cluster 3 and 7. All video clips for summarization are represented as topic profile $\gamma$. 
Recall that each local scene is learned with $K=15$ topics and scene clusters with $N_s$ scenes are learned with $K=coeff\times N_{s}$ topics where $coeff$ is set to 5 here. For fair comparison, flat model baselines are learned with the sum of the number of topics for each cluster.

\vspace{0.2cm}\noindent\textbf{Summarization Evaluation}
The performance is evaluated by the coverage of identified behaviours
in the summary, averaged over 50 independent
runs. Fig.~\ref{fig:summaryResult}(a) and (b) show the results for
multi-scene summarization within two example clusters (\textbf{Condition WC}). Clearly our
Multi-Scene Kcenter algorithm (red) outperforms the baselines: both
Graph Method alternative (purple), and single-scene alternatives
(dashed line). The performance margin is greater between multi-scene
and single-scene models for the first cluster because there are four
scenes here, so greater opportunity to exploit inter-scene
redundancy. This validates the effectiveness of jointly exploiting
multiple-scenes for summarization. Fig.~\ref{fig:summaryResult}(c)
shows the result for multi-scene summarization across both clusters (\textbf{Condition AC}):
our \textbf{Scene Cluster Model} builds one summary for each cluster to exploit the
expected greater volume of within-cluster redundancy. In contrast, the
\textbf{Flat Model} builds one single summary, but for a much more diverse
group of data, and the single-scene models have no across-cluster
redundancy to exploit. Even in the flat case, our Kcenter model
(in green) still outperforms all other alternatives (purple and
magenta). It is also worth noting that the user attention model degenerates severely on our dataset due to the inability to extract semantic meaning from videos where pure motion strength is not informative enough to distinguish semantic behaviours.
Qualitative results for multi-scene summarization are presented in supplementary material.

\begin{figure}[!ht]
\begin{center}
	\subfloat[Condition WC: Scene Cluster 3 (4 scenes total)]{\includegraphics[width = 0.95\linewidth]{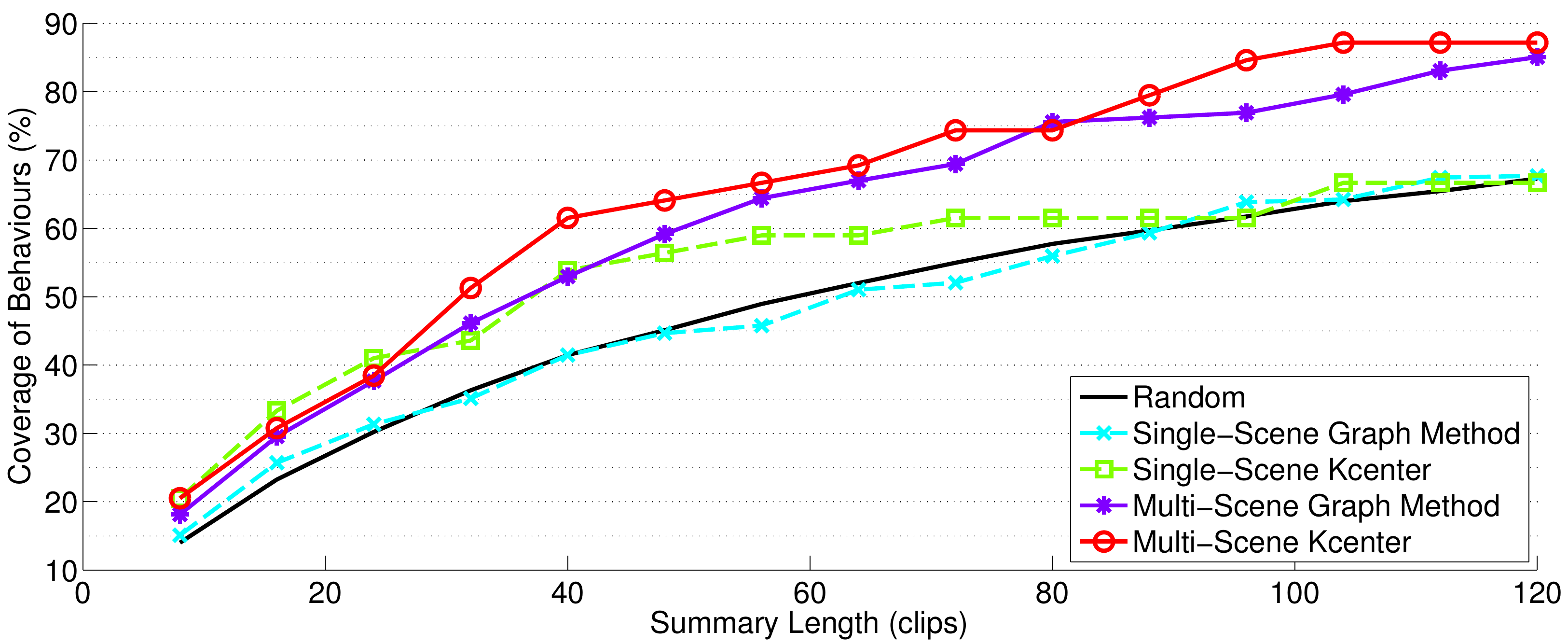}}
	
	\subfloat[Condition WC: Scene Cluster 7 (2 scenes total)]{\includegraphics[width = 0.95\linewidth]{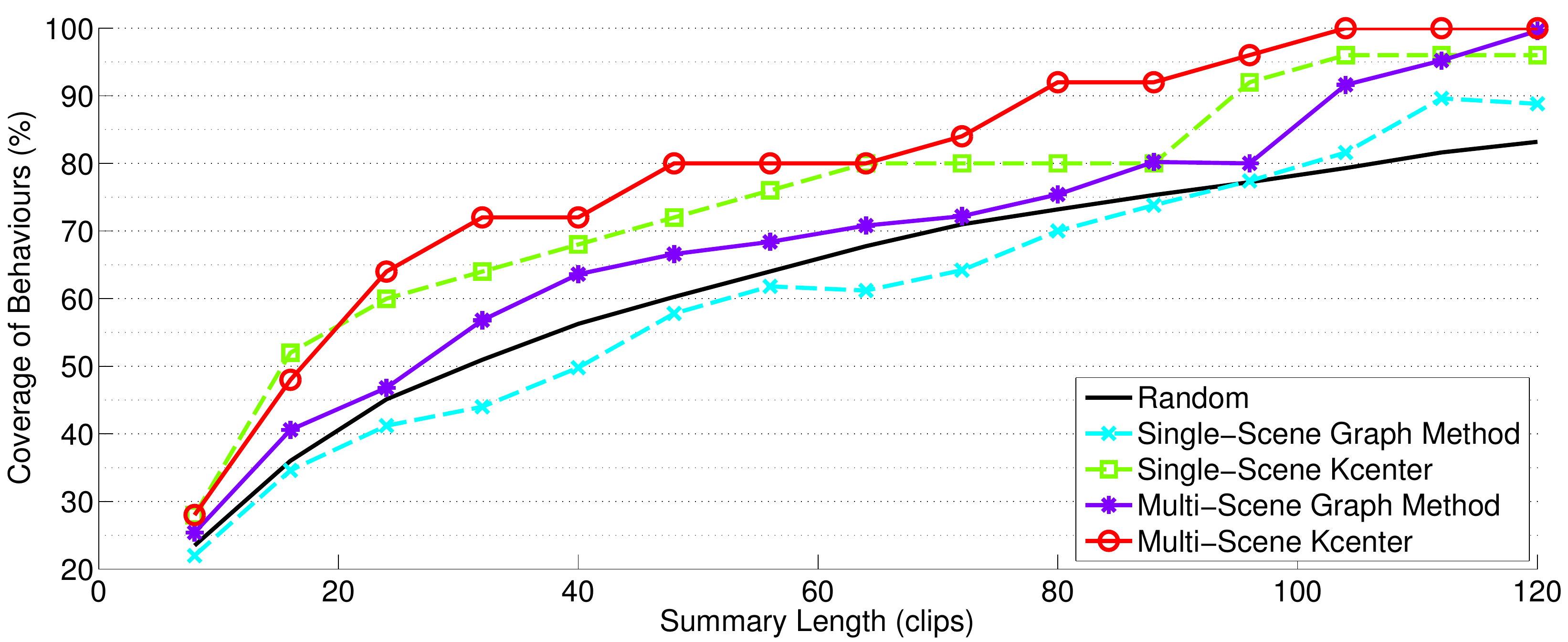}}
	
	\subfloat[Condition AC: All Scenes (6 scenes total)]{\includegraphics[width = 0.95\linewidth]{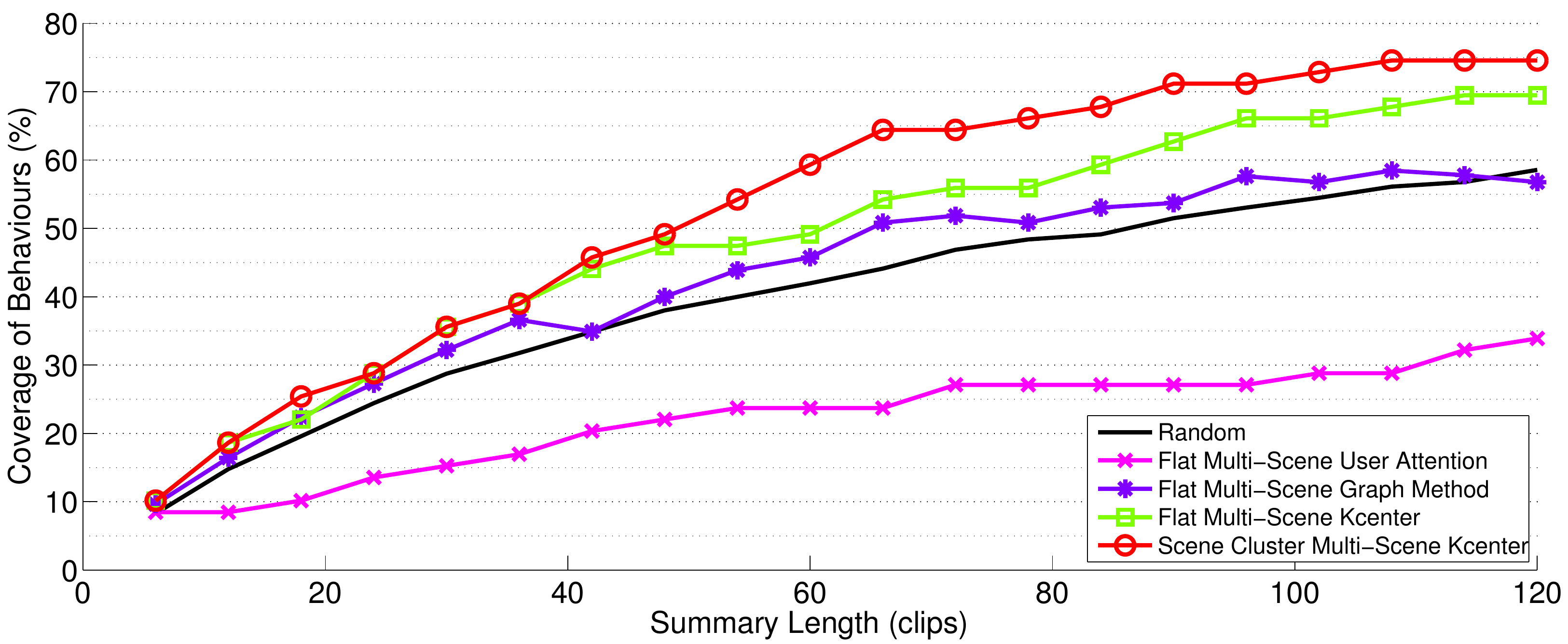}}	
	
\caption{Video summarization results: Coverage of  behaviours versus summary clip length.}
\label{fig:summaryResult}
\end{center}
\end{figure}

\vspace{-0.3cm}
{
\subsection{Further Analysis}
In this section, we further analyse the robustness of our framework,
by varying key parameters, and investigate their impact on model
performance.}\\~\vspace{-0.1cm}
\added{
\noindent{\textbf{Generalised Scene Alignment}}\quad We assume
currently that cameras are installed upright and
only scaling and translational transform are applied to scene alignment.
However, under more generally, rotational transforms
may also be considered. To that end, one can consider a generalised
scene alignment that includes a rotational parameter $\phi$ in the
transformation. Recall that in
section~\ref{subsection:SceneLevelClustering}, we estimate the
size of transformed topics. We can extend that to 
$N_a^{\prime}=N_a\times t_s \times cos(\phi)$ and
$N_b^{\prime}=N_b\times t_s \times sin(\phi)$. The generalised
transform matrix $\mathbf{T}$ is then defined as:
\begin{equation}\label{eq:ExtTransform}
\resizebox{.6\hsize}{!}
{$
\mathbf{T} = 
\begin{bmatrix}
t_s \cdot cos(\phi) & -t_s \cdot sin(\phi) & t_{x}\\
t_s \cdot sin(\phi) & t_s \cdot cos(\phi) & t_{y}\\
0 & 0 & 1
\end{bmatrix}
$}
\end{equation}
The procedure to transform a topic under this generalised alignment
differs from the original alignment only in the estimation of
direction $d$. To determine $d$ given $d^{\prime}$, we represent
quantized optical flow as vector $vec^{\prime}=[cos(2\pi d^{\prime}/N_{m}), \,\, 
     sin(2\pi d^{\prime}/N_{m})]^{\mathbf{T}}$. Then we estimate the
   original flow vector $vec=\mathbf{T}^{*-1}vec^{\prime}$ where
   $\mathbf{T}^{*}$ is a $2\times2$ matrix from the first two dimensions
   of $\mathbf{T}$ because translation does not change motion
   direction. We determine $d$ by nearest neighbour as follows:
   \begin{equation}\label{eq:Determine_d}
\hat{d}=\underset{d=1\cdots N_{m}}{argmin}\left \| vec-
\begin{bmatrix}
cos(2\pi d/N_{m})\\
sin(2\pi d/N_{m})
\end{bmatrix}\right \|
\end{equation}
To align scene A to scene B with this generalised alignment, we can
estimate parameters by maximizing the marginal likelihood of target
document $\mathbf{X_b}$ given source topics $\beta_a$. Specifically,
we denote the transform operation with specified parameters as
$\mathbb{H}(\beta|t_s,t_x,t_y,\phi)$. Given target document
$\mathbf{X_b}$, the marginal likelihood is
$p(\mathbf{X_b}|\alpha_a,\mathbb{H}(\beta_a|t_s,t_x,t_y,\phi))$ where
$\alpha_a$ is the Dirichlet prior in scene A. Because scaling and
translational parameters are computed by a closed-form solution
(Eq.~(\ref{eq:TransformParam})), we only need to search
$\hat{\phi}=\underset{\phi}{argmax}~
p(\mathbf{X_b}|\alpha_a,H(\beta_a|s,dx,dy,\phi))$. However, 
in our experiments with applying this generalised alignment process, 
we observed many local minima -- suggesting that the rotational
transform is under-constrained, and not very repeatable.
}

{
\vspace{0.1cm}\noindent{\textbf{Scene Alignment Stability}}\quad We
first evaluate the stability of scene-level alignment. Recall that
given two scenes $a$ and $b$, we firstly normalize each scene with
geometrical transformation $\mathbf{T}_{norm}^a$ and
$\mathbf{T}_{norm}^b$. The scene $a$ to $b$ transform is thus defined
by:
\begin{equation}\label{eq:SceneTformExpand}
\resizebox{.8\hsize}{!}
{$
\begin{split}
& \mathbf{T}^{a2b}=\mathbf{T}_{norm}^{b -1} \cdot \mathbf{T}_{norm}^{a}= 
\begin{bmatrix}
\frac{t^a_s}{t^b_s} & 0 & \frac{t^a_x}{t^b_s}-\frac{t^b_x}{t^b_s}\\
0 & \frac{t^a_s}{t^b_s} & \frac{t^a_y}{t^b_s}-\frac{t^b_y}{t^b_s}\\
0 & 0 & 1
\end{bmatrix}
\end{split}$}
\end{equation}
We denote $s^{a2b}=\frac{t^a_s}{t^b_s}$,
$dx^{a2b}=\frac{t^a_x}{t^b_s}-\frac{t^b_x}{t^b_s}$,
$dy^{a2b}=\frac{t^a_y}{t^b_s}-\frac{t^b_y}{t^b_s}$. The parameters
estimated from full data in each scene are denoted as $s^{a2b}_{ref}$,
$dx^{a2b}_{ref}$, $dy^{a2b}_{ref}$. To evaluate the stability of this
alignment, we randomly sample $50\%$ of the original data from each
scene and estimate again the parameters as $s^{a2b}_{50}$,
$dx^{a2b}_{50}$, $dy^{a2b}_{50}$.  We run this process for 20 times
and calculate the Root Mean Square Error (RMSE), defined in
Eq.~(\ref{eq:CVRMSE}) for $s^{a2b}$. RMSE for $dx$ and $dy$ are
defined in the same way by replacing $s^{a2b}$ with $dx^{a2b}$ and
$dy^{a2b}$ respectively. 
\begin{equation}\label{eq:CVRMSE}
\resizebox{.6\hsize}{!}
{$RMSE(s) = {\sqrt{\frac{1}{N}\sum\limits_{i=1}^{N} (s^{a2b}_{50 i} - s^{a2b}_{ref})^2}}$}
\end{equation}

We show both the absolute value of reference parameters and RMSE when aligning each pair of scenes in Fig.\ref{Fig:CVRMSE_ScnPairs}.
\begin{figure}[t]
\vspace{-0.5cm}
\begin{center}
\subfloat[Absolute reference value of scaling]{\includegraphics[width=0.3\linewidth]{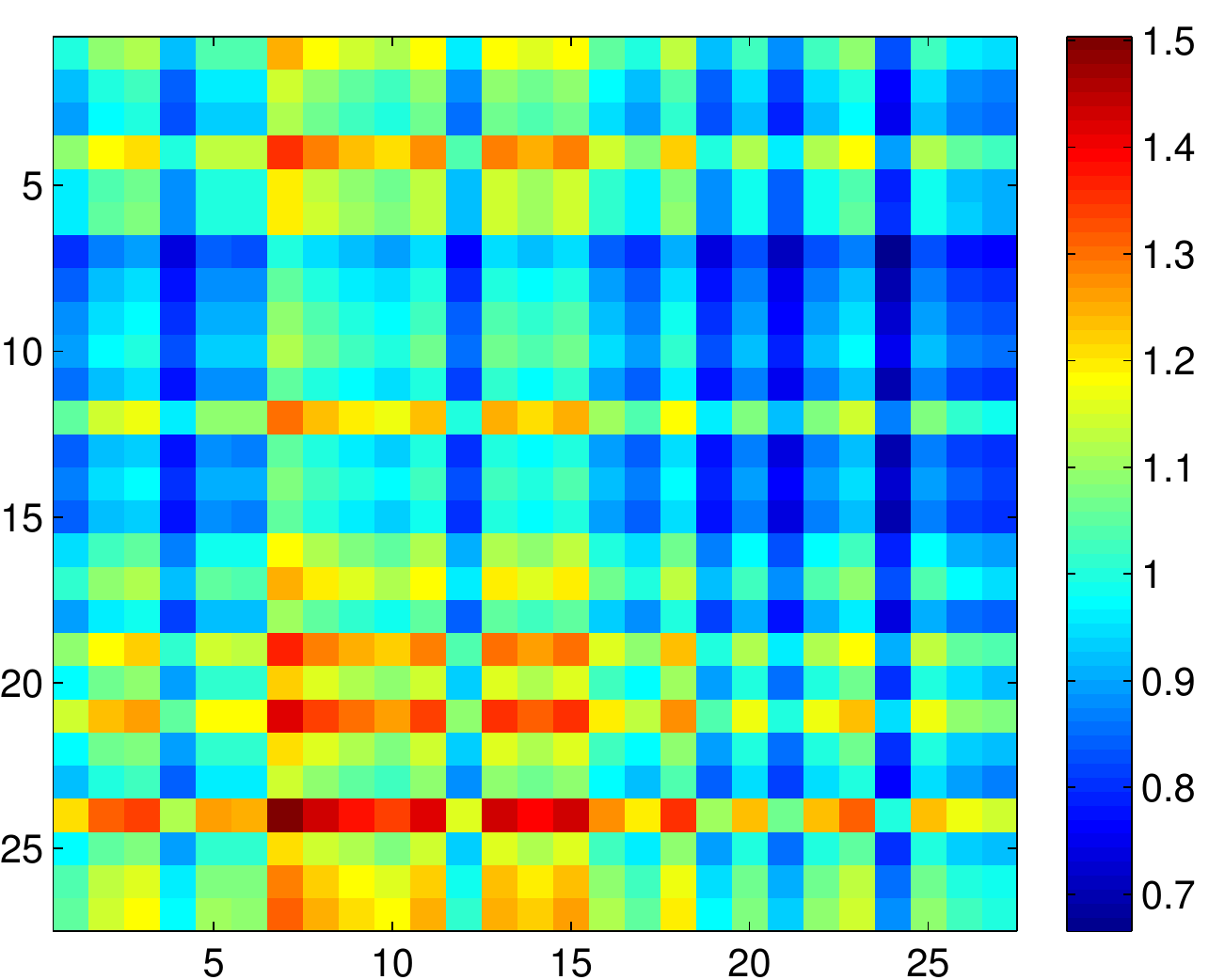}}\hspace{0.01cm}
\subfloat[Absolute reference value of x translation]{\includegraphics[width=0.3\linewidth]{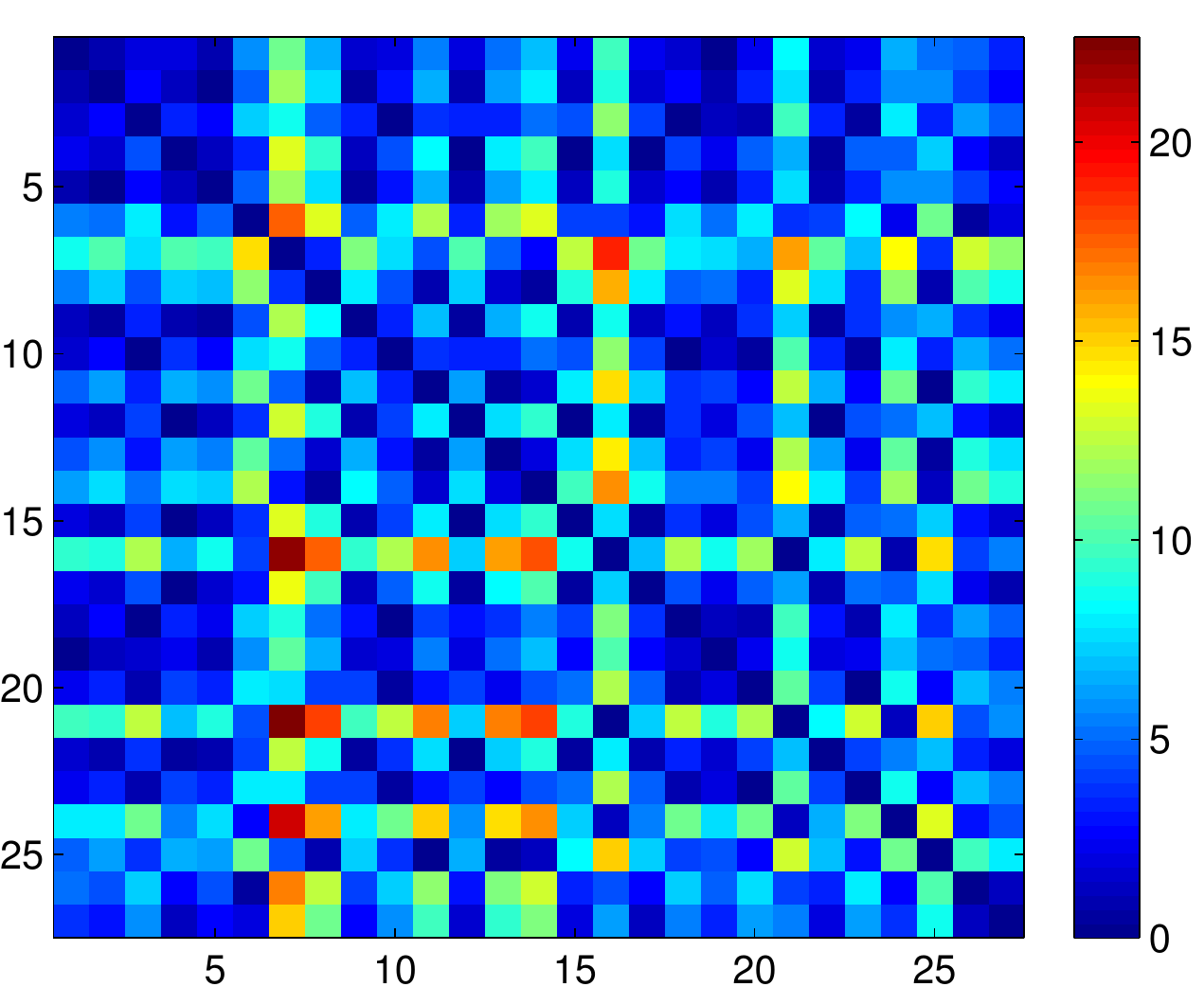}}\hspace{0.01cm}
\subfloat[Absolute reference value of y translation]{\includegraphics[width=0.3\linewidth]{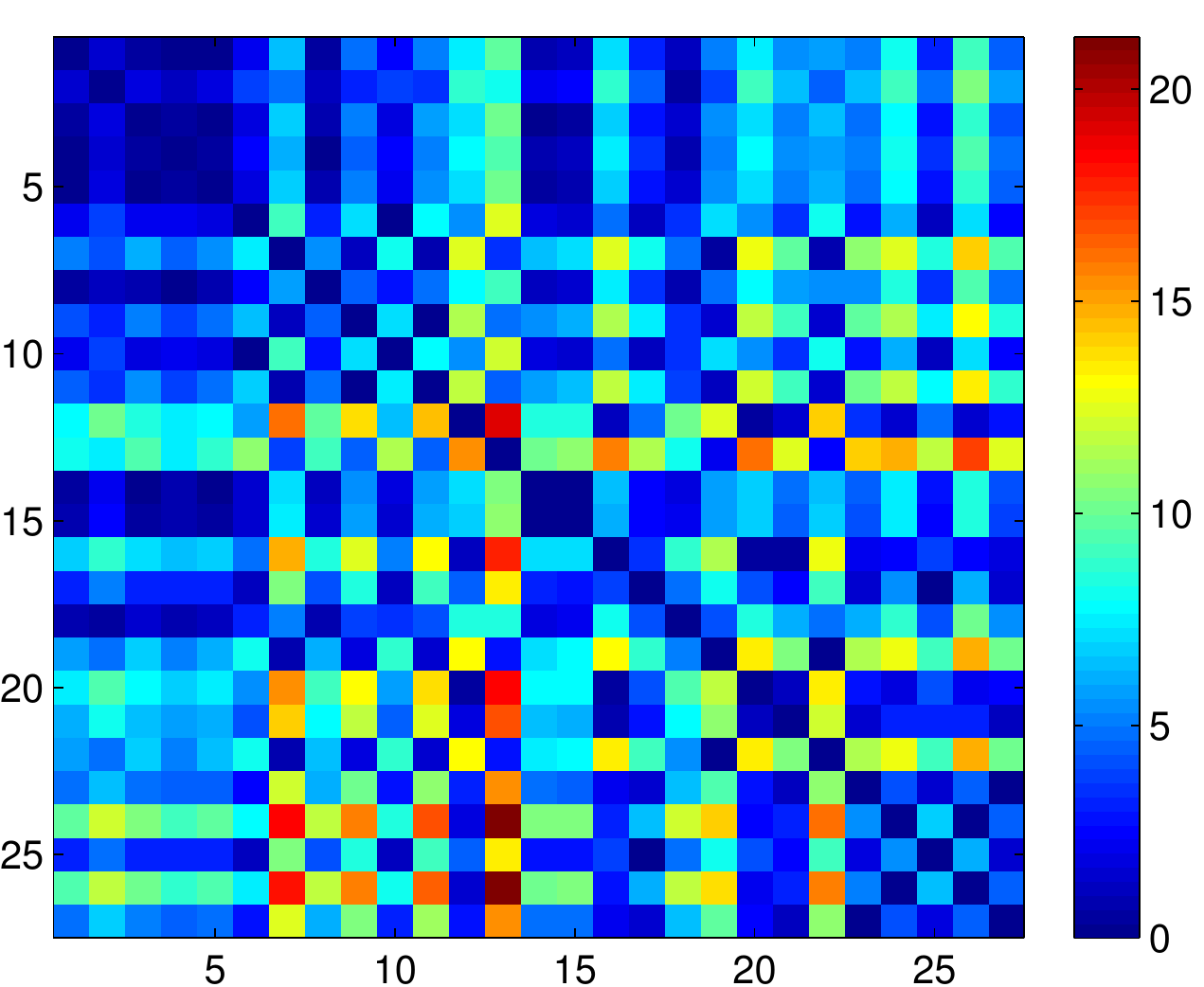}}\\
\subfloat[RMSE of scaling]{\includegraphics[width=0.3\linewidth]{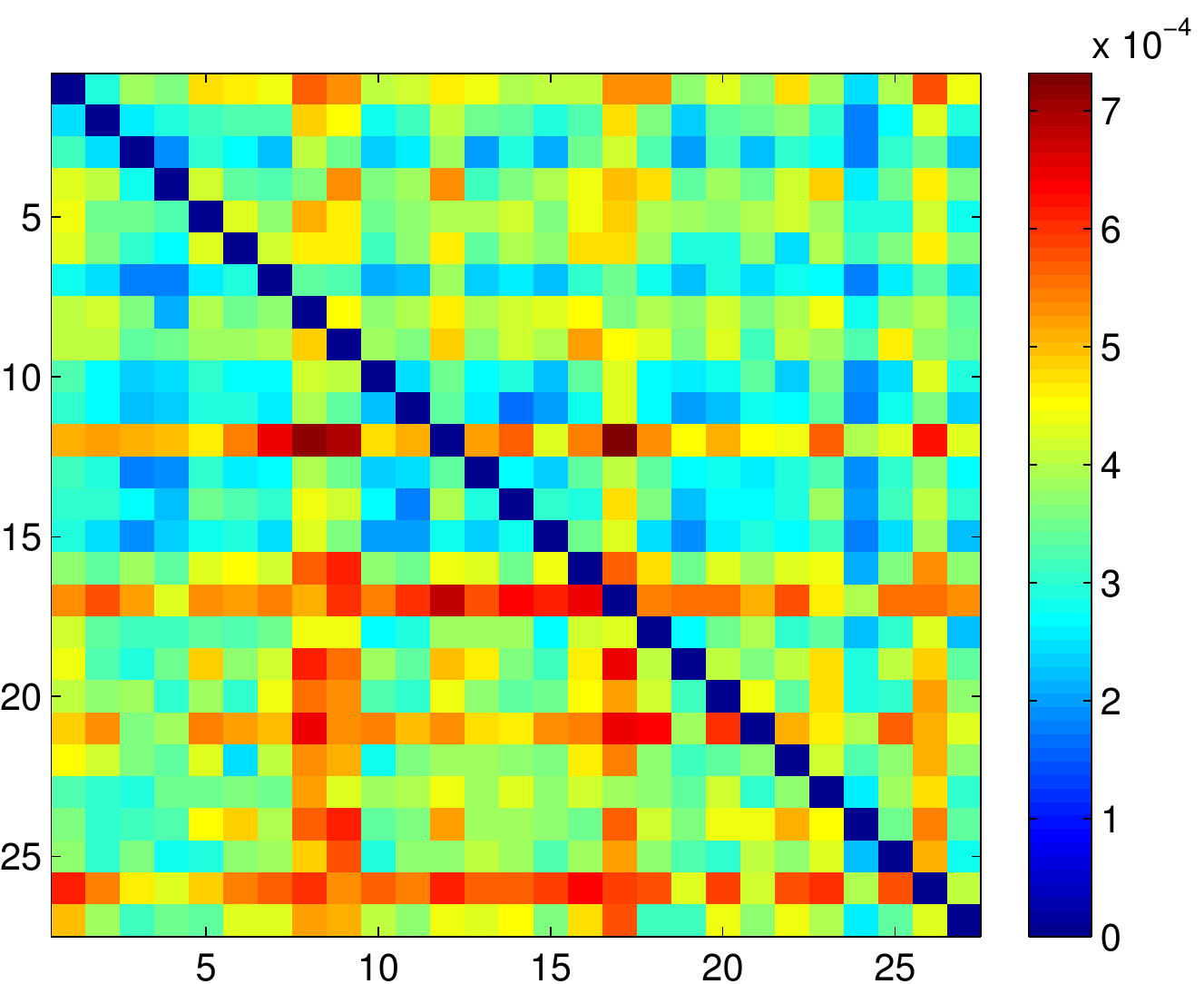}}\hspace{0.01cm}
\subfloat[RMSE of x translation]{\includegraphics[width=0.3\linewidth]{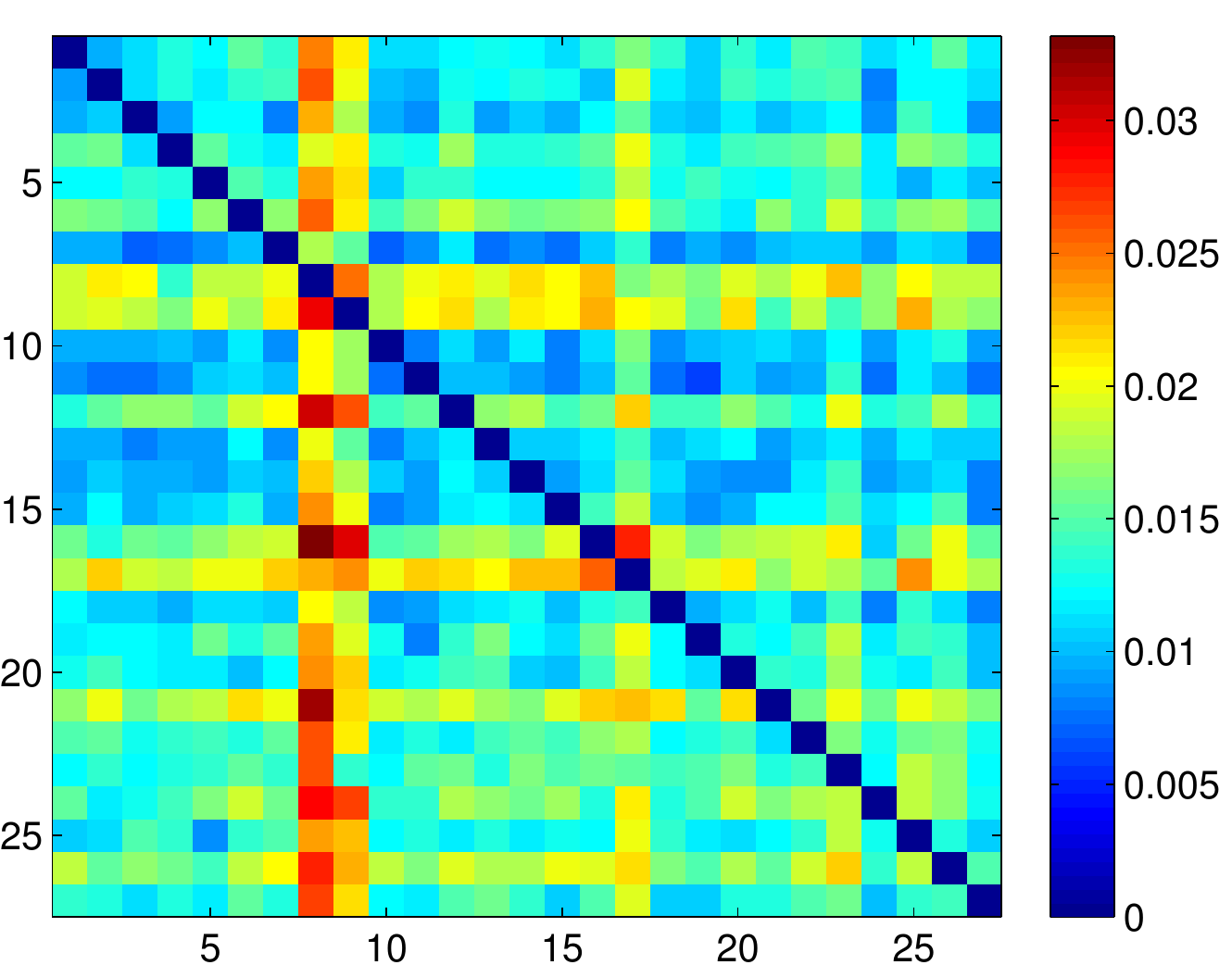}}\hspace{0.01cm}
\subfloat[RMSE of y translation]{\includegraphics[width=0.3\linewidth]{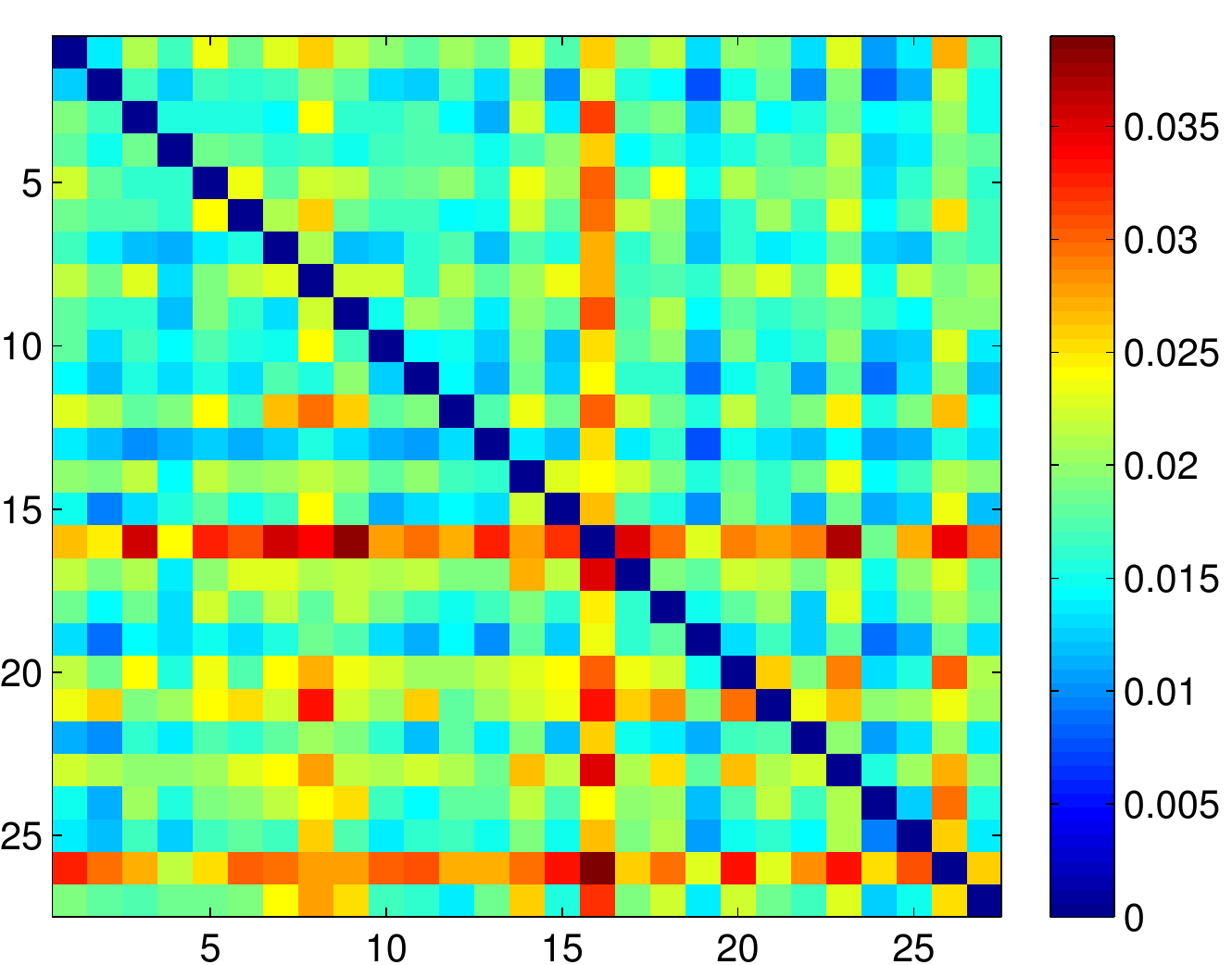}}
\caption{Alignment and stability across all pairs of 27 scenes.}\label{Fig:CVRMSE_ScnPairs}
\end{center}
\vspace{-0.5cm}
\end{figure}
\break
It is evident that most scene pairs are scaled between $0.7$ and $1.5$
(Fig.~\ref{Fig:CVRMSE_ScnPairs}(a)). The worst RMSE(s) among all scene
pairs is $0.0007$ (Fig.~\ref{Fig:CVRMSE_ScnPairs}(d)). The same
observations can be made on variability of x translation and y
translation with the largest RMSE(dx) and RMSE(dy) being $0.035$ pixels or
less while the absolute value of reference x and y translation are
between $0$ and $20$ pixels. The small values of these deviations
verify that the scene alignment model is robust and repeatable.
\added{
Some examples of scene alignment are shown in
Fig.~\ref{fig:ExampleAlign}. Whilst the majority of
activities are aligned well, some are less so. This is
due to the limitation of a global rigid transform over a whole
scene. Further extension could exploit individual
activity centered alignment in addition to holistic scene alignment.
\begin{figure}
\centering
{\includegraphics[width = 0.24\linewidth]{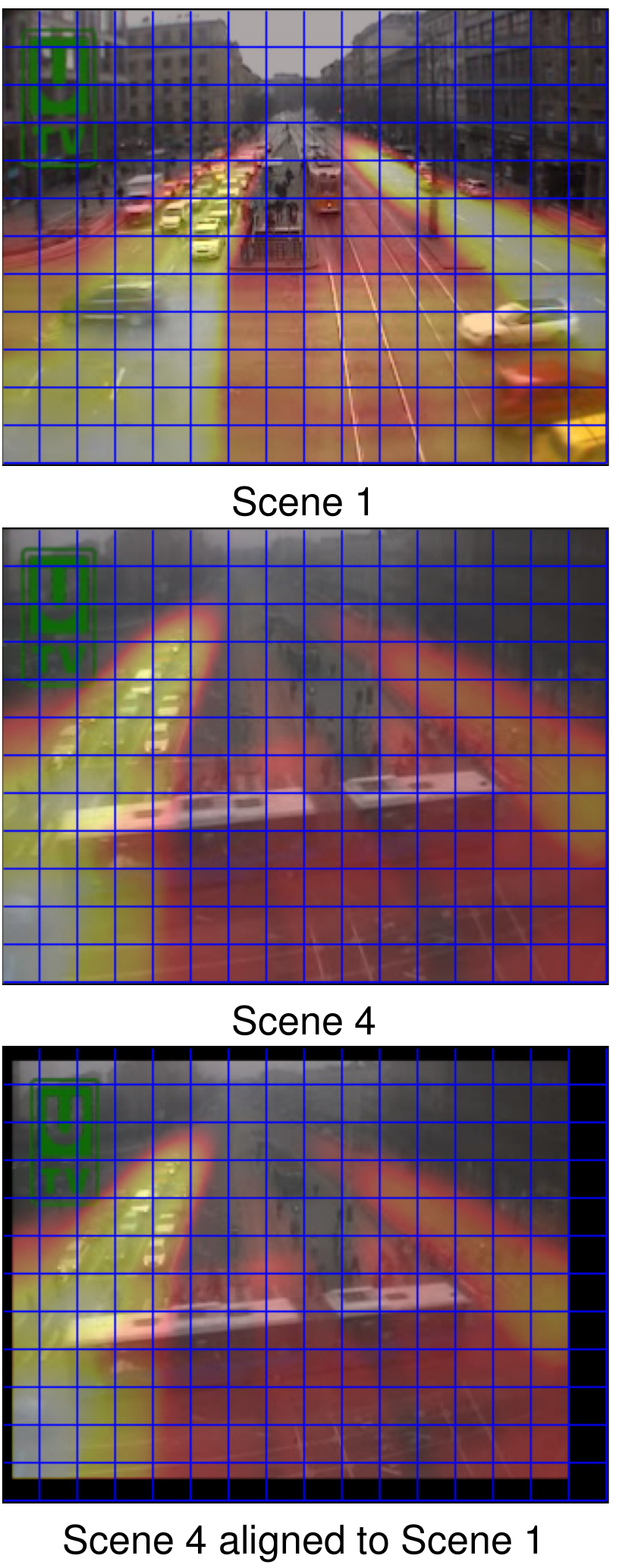}}
{\includegraphics[width = 0.24\linewidth]{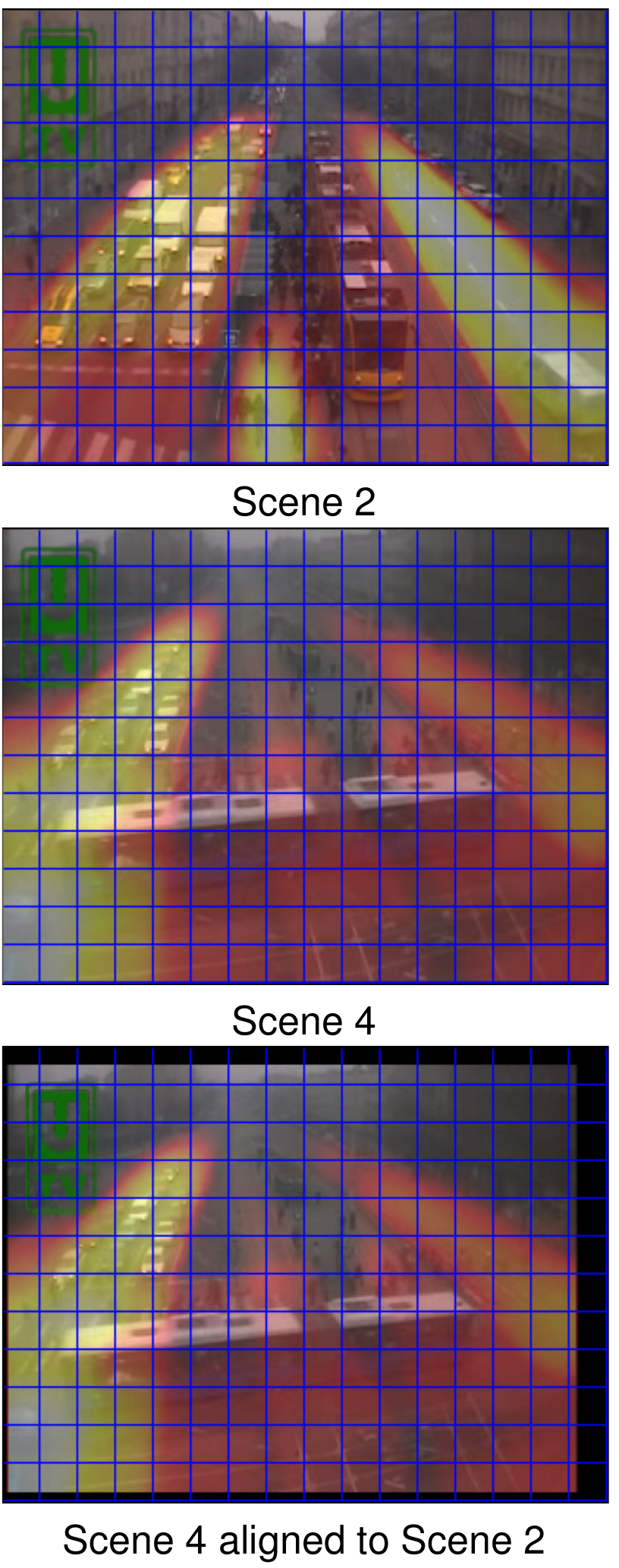}}
{\includegraphics[width = 0.24\linewidth]{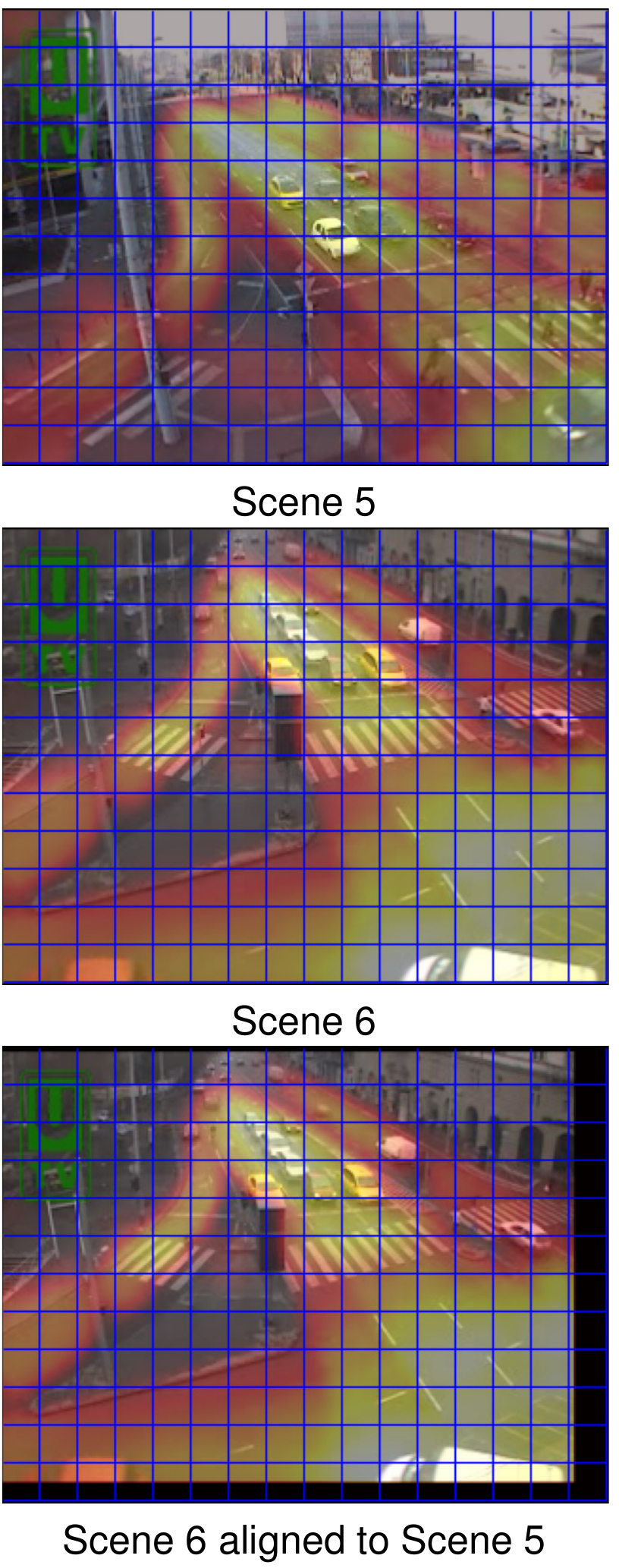}}
{\includegraphics[width = 0.24\linewidth]{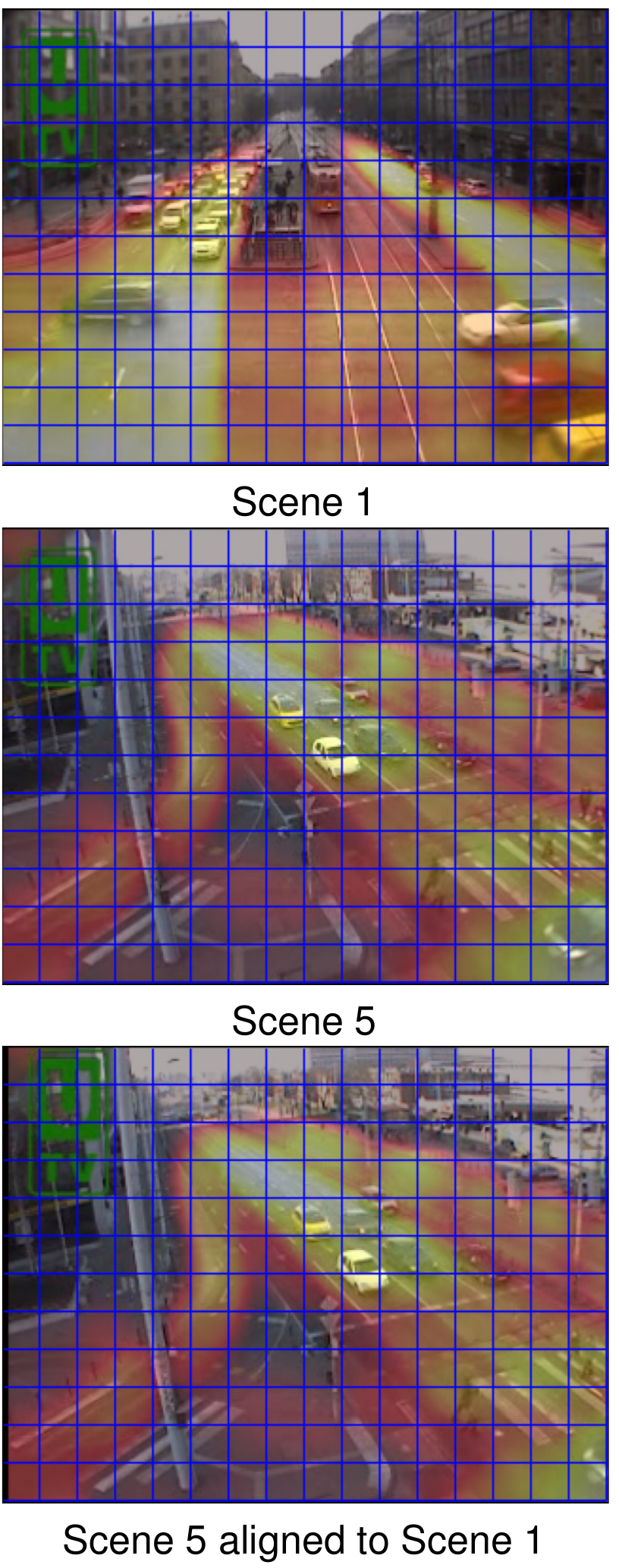}}
\caption{Examples of scene alignment pairs. Each column indicates one example alignment. The first row is the target scene, the second row is the source scene to be aligned/transformed and the last row is the source scene after alignment to the target. Both within scene cluster (first three
  columns, clusters 3, 3 and 7 respectively) and across cluster (fourth column, cluster 3 and 7) examples are presented. The overlaid heat map is the spatial frequency of visual  words.}\label{fig:ExampleAlign}
\end{figure}
}
}

{
\vspace{0.1cm}\noindent{\textbf{Scene Cluster Stability}}\quad
We tested the stability of scene-level clustering by varying cell
size, number of local topics, and clustering strategy: (1) We compared
visual word quantisation with $5 \times 5$ and $10 \times 10$ cell
size. (2) We evaluated from 5 to 30 local topics in each scene by step
of 5. (3) We performed self-tuning spectral clustering with two
alternative settings. The first is that we allowed the model to automatically
determine number of clusters and the second is that we fixed the number of
clusters to the same as in the reference clustering, that is, 15 local
topics and $5 \times 5$ cell size. We measured the discrepancy between
the results from automatic clustering and the reference clustering using
the Rand Index \cite{rand1971}. It describes the discrepancy between two
set partitions and is frequently used as the evaluation metric for
clustering. The Rand Index is between 0 and 1, with the higher value
indicating more similar between two partitions. If two partitions are
exactly the same, the Rand Index is 1. We show the results on the
stability test of scene-level clustering in
Fig.\ref{fig:SceneClusterStability}.
\begin{figure}[!h]
\vspace{-0.5cm}
\begin{center}
\subfloat[Rand Index(RI) cell size=5]{\includegraphics[width=0.56\linewidth]{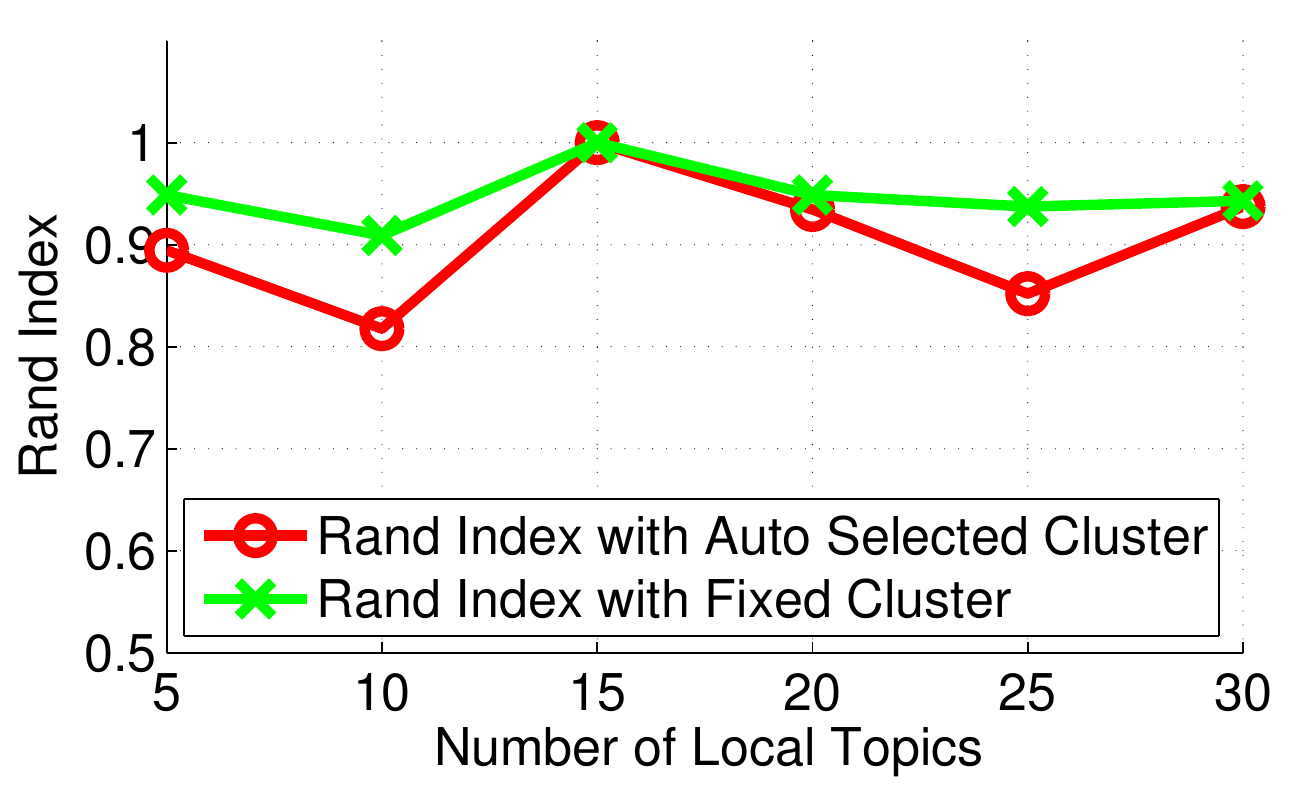}}\vspace{-0.4cm}
\subfloat[Number of cluster cell size=5]{\includegraphics[width=0.38\linewidth]{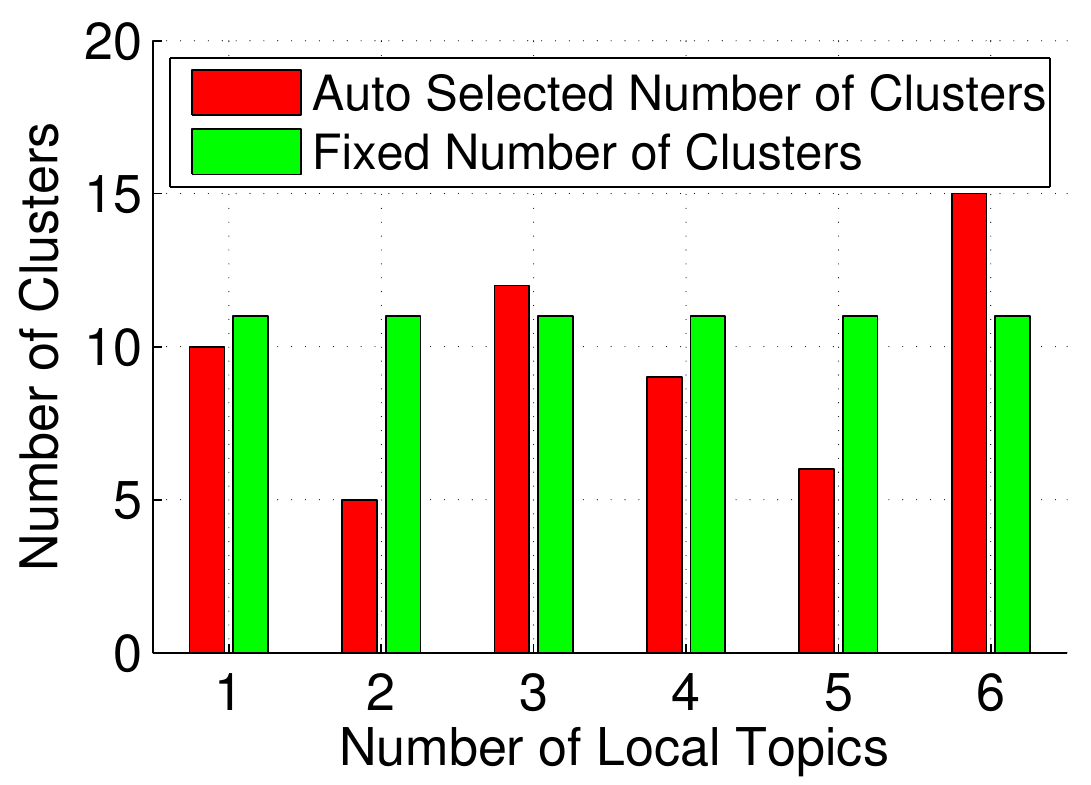}}\vspace{-0.4cm}\\
\subfloat[Rand Index(RI) cell size=10]{\includegraphics[width=0.56\linewidth]{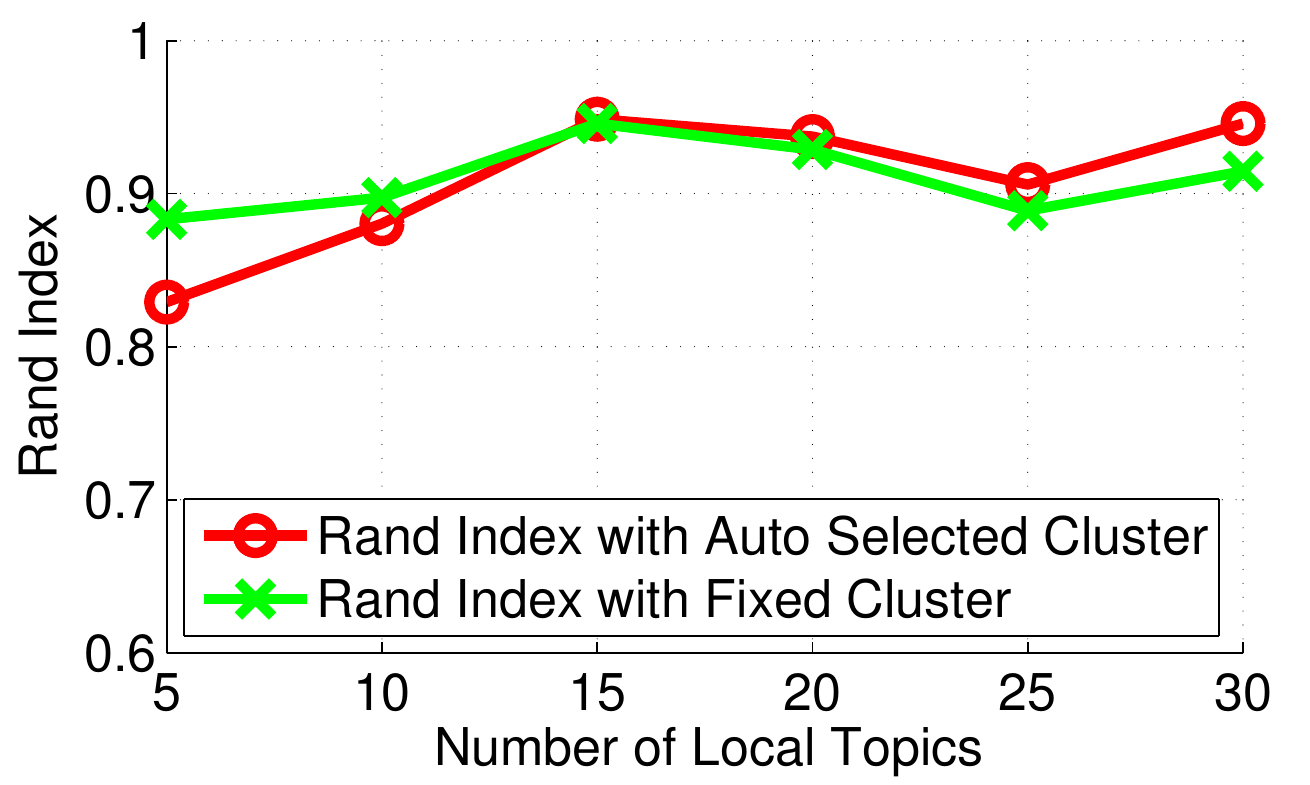}}
\subfloat[Number of cluster cell size=10]{\includegraphics[width=0.38\linewidth]{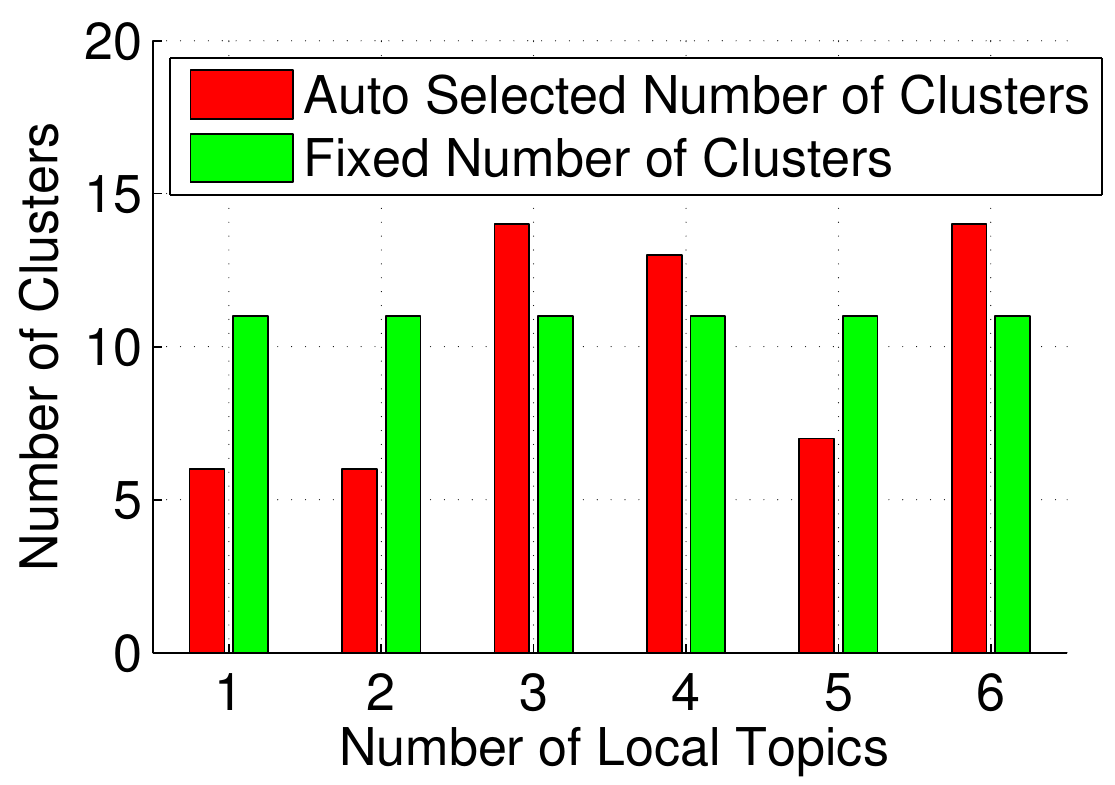}}
\caption{Stability of scene-level clustering.
}\label{fig:SceneClusterStability}
\end{center}
\vspace{-0.4cm}
\end{figure}
}

{
For both cell size = 5 and = 10, automatic cluster
selection generates consistent partitions (high Rand
Index). So the framework is robust to motion quantisation cell size.
However, it is also evident that automatic cluster number selection is
less stable in determining the number of clusters as indicated by
the red bars in Fig.\ref{fig:SceneClusterStability}(b) and (d).
On the other hand, by fixing the number of clusters, 
the partitioning is more stable (consistent high Rand Index).}

{
\vspace{0.1cm}\noindent{\textbf{Associating New Scenes}}\quad
Our model is able to group scenes according to the semantic
relatedness if all the recorded data are available in
advance. In addition, the model is capable of associating new scenes
to existing clusters, e.g. given input from newly installed cameras at
different locations, without the need to completely re-learn the model.
This is achieved by comparing the local topics of a new scene to the
STB in each scene cluster and choosing the cluster with highest
relatedness. Only the updated cluster needs to be re-learned to
incorporate the new scene. We tested this approach in Scene Clusters 3
and 7 by: (1) Hold out each scene in turn as the candidate scene to be
associated and learn STB in each cluster with the other scenes; (2)
compute the relatedness between the held-out scene and both clusters
using Eq.~(\ref{eq:SceneSimilarity}); (3) associate the candidate
scene to the cluster with the highest relatedness. We illustrate the
result of this via the distance (defined as $1-$ relatedness) between
held-out candidate scenes and clusters in
Fig.~\ref{Fig:SubsetAssocScene}. It is evident that each held out
scene is closer to its corresponding cluster, so 100\% of scenes are
associated correctly. However, this approach is limited to associating
new scenes to existing scene clusters (scenes). A full online learning
multi-scene model is desirable but also challenging and remains to
be developed. }

\begin{figure}
\begin{center}
\includegraphics[width=0.65\linewidth]{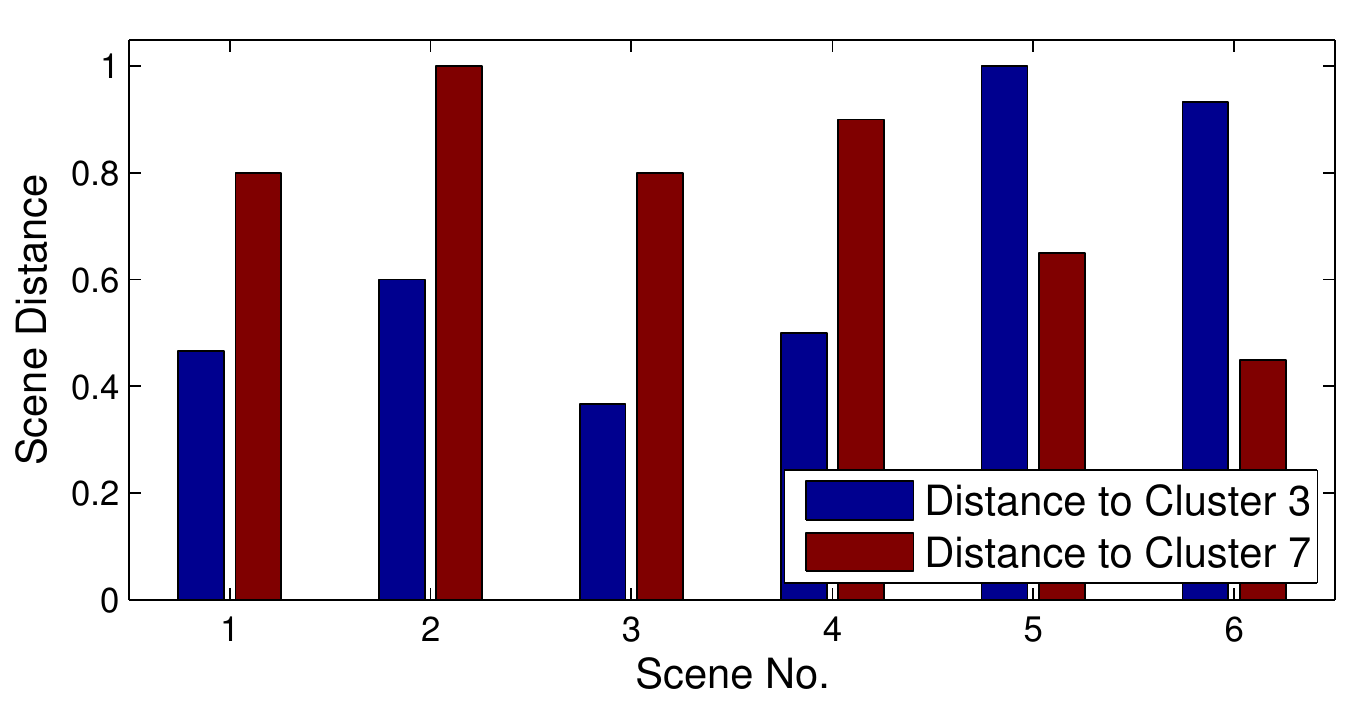}
\caption{Association of held out-scenes to clusters. Scene 1-4 are held out from cluster 3, and scene 5-6 are held-out scenes from cluster 7. All held-out scenes are correctly associated.}\label{Fig:SubsetAssocScene}
\end{center}
\vspace{-0.5cm}
\end{figure}

{
\vspace{0.1cm}\noindent{\textbf{STB Stability}}\quad
Finally, we investigate the stability of learning the Shared Topic
Basis (STB) with different number of shared topics. Recall that, in
section \ref{subsection:SceneLevelClusteringResult}, the number of STB
topics for the Scene Cluster Model (SCM) and the Flat Model (FM) is $K=coeff
\times N_s$. Now let us change $coeff$ from 3 to 10 and evaluate how  this
affects the cross-scene classification accuracy for both annotation
Scheme 1 (59 categories) and 2 (31 categories). The results are shown
in Fig.~\ref{fig:STB_Stability_Classification}. 
\begin{figure}[h]
\vspace{-0.3cm}
\begin{center}
\includegraphics[width=0.8\linewidth]{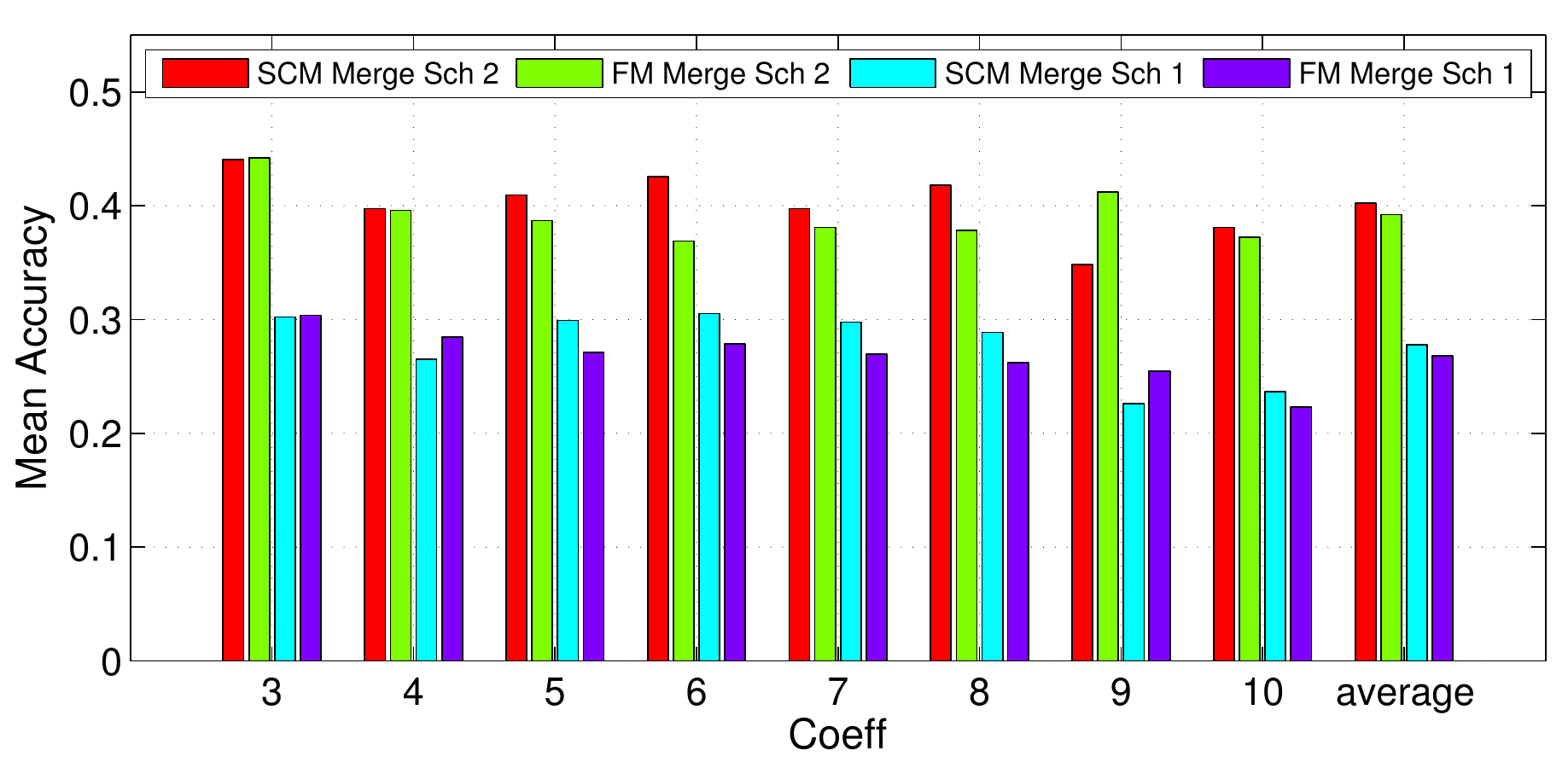}
\caption{Effect of varying number of topics used. Classification accuracy of Scene Cluster Model (SCM) and Flat Model (FM).}\label{fig:STB_Stability_Classification}
\end{center}
\end{figure}
It is evident that for both 59 and 31 categories, our Scene Cluster
Model is mostly better than Flat Model over a range of topic
numbers.
}

\section{Conclusions}

In this paper we introduced a framework for synergistically modelling multiple-scene datasets captured by multi-camera surveillance networks. It deals with variable and piece-wise inter-scene relatedness by semantically clustering scenes according to the correspondence of semantic activities; and selectively shares activities across scenes within clusters. Besides revealing the commonality and uniqueness of each scene, multi-scene profiling further enables typical surveillance tasks of query-by-example, behaviour classification and summarization to be generalised to multiple scenes. Importantly, by discovering related scenes and shared activities, it is possible to achieve cross-scene query-by-example  (in contrast to typical within-scene query), and to annotate behaviour in a novel scene without any labels -- which is important for making deployment of surveillance systems scale in practice. Finally, we can provide video summarization capabilities that  uniquely exploit redundancy both within and across scenes by leveraging our multi-scene model.

There are still several limitations to our work which can be addressed in the future: (i) In the current framework, scenes that can be grouped together are usually morphologically similar, which means the underlying motion patterns and view angles are essentially similar. More advanced geometrical registration techniques could be applied, including similarity and affine transformations, to allow scenes with more dramatic viewpoint changed to be grouped. (ii)  In this work motion information is mostly contributed by traffic. However studying pedestrian/crowd behaviour is becoming more interesting \cite{Shao_2014_CVPR} due to  wide application in crime prevention and public security. However, compared with traffic, pedestrian crowd behaviours are less regulated and coherent. Thus, exacting suitable features and improving the model to deal with this are non-trivial tasks. (iii) Finally, an improved multi-scene framework that can fully incrementally add new scenes in an online manner is of interest.


%

%
%
%
%
%

\ifCLASSOPTIONcaptionsoff
  \newpage
\fi

\begin{IEEEbiography}[{\includegraphics[width=1in,height=1.25in,clip,keepaspectratio]{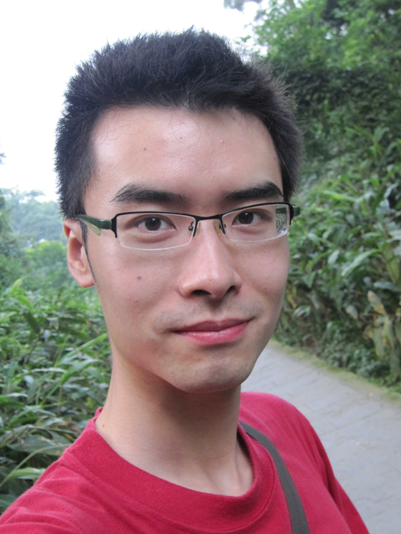}}]{Xun Xu} is a Ph.D. candidate in the School of Electronic Engineering and Computer Science, Queen Mary, University of London, UK. He received his B.E. degree in the School of Electrical Engineering and Information, Sichuan University, Chengdu, in 2010. His research interests include surveillance video understanding, transfer learning and event recognition.
\end{IEEEbiography}

\begin{IEEEbiography}[{\includegraphics[width=1in,height=1.25in,clip,keepaspectratio]{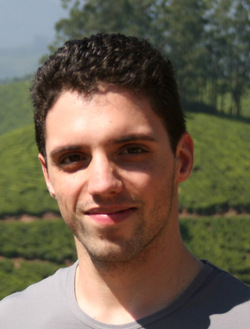}}]{Timothy Hospedales} received the Ph.D degree in Neuroinformatics from University of Edinburgh in 2008 and now is a Lecturer (Assistant Professor) in Computer Science at Queen Mary University of London. His research interests include transfer and multi-task machine learning applied to problems in computer vision and beyond. He has published over 30 papers in major international journals and conferences.
\end{IEEEbiography}

\begin{IEEEbiography}[{\includegraphics[width=1in,height=1.25in,clip,keepaspectratio]{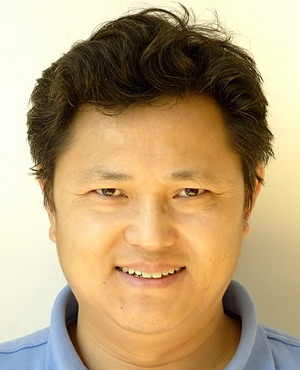}}]{Shaogang
    Gong} is Professor of Visual Computation at Queen Mary University
  of London, a Fellow of the Institution of Electrical Engineers and a
  Fellow of the British Computer Society. He received his D.Phil in
  computer vision from Keble College, Oxford University in 1989. His
  research interests include computer vision, machine learning and
  video semantic analysis. \end{IEEEbiography}






\begin{thebibliography}{10}
\providecommand{\url}[1]{#1}
\csname url@samestyle\endcsname
\providecommand{\newblock}{\relax}
\providecommand{\bibinfo}[2]{#2}
\providecommand{\BIBentrySTDinterwordspacing}{\spaceskip=0pt\relax}
\providecommand{\BIBentryALTinterwordstretchfactor}{4}
\providecommand{\BIBentryALTinterwordspacing}{\spaceskip=\fontdimen2\font plus
\BIBentryALTinterwordstretchfactor\fontdimen3\font minus
  \fontdimen4\font\relax}
\providecommand{\BIBforeignlanguage}[2]{{%
\expandafter\ifx\csname l@#1\endcsname\relax
\typeout{** WARNING: IEEEtran.bst: No hyphenation pattern has been}%
\typeout{** loaded for the language `#1'. Using the pattern for}%
\typeout{** the default language instead.}%
\else
\language=\csname l@#1\endcsname
\fi
#2}}
\providecommand{\BIBdecl}{\relax}
\BIBdecl

\bibitem{journal/pami/WangMG09}
X.~Wang, X.~Ma, and W.~Grimson, ``Unsupervised activity perception in crowded
  and complicated scenes using hierarchical bayesian models,'' \emph{IEEE
  Transactions on Pattern Analysis and Machine Intelligence}, vol.~31, no.~3,
  pp. 539--555, 2009.

\bibitem{journal/IJCV/HospedalesLGX2012}
T.~Hospedales, S.~Gong, and T.~Xiang, ``Video behaviour mining using a dynamic
  topic model,'' \emph{International Journal of Computer Vision}, vol.~98, pp.
  303--323, 2012.

\bibitem{journals/pami/HuXFXTM06}
W.~Hu, X.~Xiao, Z.~Fu, D.~Xie, T.~Tan, and S.~J. Maybank, ``A system for
  learning statistical motion patterns.'' \emph{IEEE Transaction on Pattern
  Analysis and Machine Intelligence}, vol.~28, no.~9, pp. 1450--1464, 2006.

\bibitem{journals/ijcv/VaradarajanEO13}
J.~Varadarajan, R.~Emonet, and J.-M. Odobez, ``A sequential topic model for
  mining recurrent activities from long term video logs.'' \emph{International
  Journal of Computer Vision}, vol. 103, no.~1, pp. 100--126, 2013.

\bibitem{conf/cvpr/KuettelBVF10}
D.~Kuettel, M.~Breitenstein, L.~Van~Gool, and V.~Ferrari, ``What's going on?
  discovering spatio-temporal dependencies in dynamic scenes,'' in \emph{IEEE
  Conference on Computer Vision and Pattern Recognition}, 2010, pp. 1951--1958.

\bibitem{journals/pami/PritchRP08}
Y.~Pritch, A.~Rav-Acha, and S.~Peleg, ``Nonchronological video synopsis and
  indexing.'' \emph{IEEE Transactions on Pattern Analysis and Machine
  Intelligence}, vol.~30, no.~11, pp. 1971--1984, 2008.

\bibitem{journal/IKDE/PanY2010}
S.~J. Pan and Q.~Yang, ``A survey on transfer learning,'' \emph{IEEE
  Transaction on Knowledge and Data Engineering}, vol.~22, no.~10, pp.
  1345--1359, Oct. 2010.

\bibitem{Xu:2013:CTS:2510650.2510657}
X.~Xu, S.~Gong, and T.~Hospedales, ``Cross-domain traffic scene understanding
  by motion model transfer,'' in \emph{Proceedings of the 4th ACM/IEEE
  International Workshop on ARTEMIS}, 2013, pp. 77--86.

\bibitem{conf/iccv/KhokharSS11}
S.~Khokhar, I.~Saleemi, and M.~Shah, ``Similarity invariant classification of
  events by kl divergence minimization.'' in \emph{IEEE International
  Conference on Computer Vision}, 2011.

\bibitem{journals/pami/MorrisT11}
B.~Morris and M.~Trivedi, ``Trajectory learning for activity understanding:
  Unsupervised, multilevel, and long-term adaptive approach,'' \emph{IEEE
  Transactions on Pattern Analysis and Machine Intelligence}, vol.~33, no.~11,
  pp. 2287--2301, 2011.

\bibitem{journals/ijcv/WangMNG11}
X.~Wang, K.~Ma, G.-W. Ng, and W.~Grimson,
  ``\BIBforeignlanguage{English}{Trajectory analysis and semantic region
  modeling using nonparametric hierarchical bayesian models},''
  \emph{\BIBforeignlanguage{English}{International Journal of Computer
  Vision}}, vol.~95, no.~3, pp. 287--312, 2011.

\bibitem{conf/iccv/KimLE11}
K.~Kim, D.~Lee, and I.~A. Essa, ``Gaussian process regression flow for analysis
  of motion trajectories.'' in \emph{IEEE International Conference on Computer
  Vision}, 2011, pp. 1164--1171.

\bibitem{journals/tcsv/PiciarelliMF08}
C.~Piciarelli, C.~Micheloni, and G.~L. Foresti, ``Trajectory-based anomalous
  event detection.'' \emph{IEEE Transaction on Circuits and Systems for Video
  Technology}, vol.~18, no.~11, pp. 1544--1554, 2008.

\bibitem{journals/jstsp/FanaswalaK13}
M.~Fanaswala and V.~Krishnamurthy, ``Detection of anomalous trajectory patterns
  in target tracking via stochastic context-free grammars and reciprocal
  process models.'' \emph{IEEE Journal of Selected Topics in Signal
  Processing}, vol.~7, no.~1, pp. 76--90, 2013.

\bibitem{DBLP:journals/ijcv/LiGX12}
J.~Li, S.~Gong, and T.~Xiang, ``Learning behavioural context,''
  \emph{International Journal of Computer Vision}, vol.~97, no.~3, pp.
  276--304, 2012.

\bibitem{journals/pami/LoyXG12}
C.~C. Loy, T.~Xiang, and S.~Gong, ``Incremental activity modeling in multiple
  disjoint cameras.'' \emph{IEEE Transactions on Pattern Analysis and Machine
  Intelligence}, vol.~34, no.~9, pp. 1799--1813, 2012.

\bibitem{journals/pami/WangTG10}
X.~Wang, K.~Tieu, and W.~E.~L. Grimson, ``Correspondence-free activity analysis
  and scene modeling in multiple camera views.'' \emph{IEEE Transactions on
  Pattern Analysis and Machine Intelligence}, vol.~32, no.~1, pp. 56--71, 2010.

\bibitem{conf/cvpr/LiPT05}
F.-F. Li and P.~Perona, ``A bayesian hierarchical model for learning natural
  scene categories.'' in \emph{IEEE Conference on Computer Vision and Pattern
  Recognition}, 2005, pp. 524--531.

\bibitem{conf/cvpr/LiS009}
L.-J. Li, R.~Socher, and F.-F. Li, ``Towards total scene understanding:
  Classification, annotation and segmentation in an automatic framework.'' in
  \emph{IEEE Conference on Computer Vision and Pattern Recognition}, 2009, pp.
  2036--2043.

\bibitem{journal/tip/Hu07}
W.~Hu, D.~Xie, Z.~Fu, W.~Zeng, and S.~Maybank, ``Semantic-based surveillance
  video retrieval,'' \emph{IEEE Transactions on Image Processing}, vol.~16,
  no.~4, pp. 1168--1181, Apr. 2007.

\bibitem{XiangGong:PR08}
T.~Xiang and S.~Gong, ``Activity based surveillance video content modelling,''
  \emph{Pattern Recognition}, vol.~41, no.~7, pp. 2309--2326, 2008.

\bibitem{journals/tomccap/TruongV07}
B.~T. Truong and S.~Venkatesh, ``Video abstraction: A systematic review and
  classification.'' \emph{TOMCCAP}, vol.~3, no.~1, 2007.

\bibitem{journals/prl/AvilaLLA11}
S.~E.~F. de~Avila, A.~P.~B. Lopes, A.~da~Luz~Jr., and
  A.~de~Albuquerque~Araújo, ``Vsumm: A mechanism designed to produce static
  video summaries and a novel evaluation method.'' \emph{Pattern Recognition
  Letters}, vol.~32, no.~1, pp. 56--68, 2011.

\bibitem{conf/iccv/PritchRGP07}
Y.~Pritch, A.~Rav-Acha, A.~Gutman, and S.~Peleg, ``Webcam synopsis: Peeking
  around the world.'' in \emph{IEEE International Conference on Computer
  Vision}, 2007, pp. 1--8.

\bibitem{conf/mm/LouCL05}
J.-G. Lou, H.~Cai, and J.~Li, ``A real-time interactive multi-view video
  system.'' in \emph{ACM Multimedia}, 2005, pp. 161--170.

\bibitem{journals/tmm/FuGZLSZ10}
Y.~Fu, Y.~Guo, Y.~Zhu, F.~Liu, C.~Song, and Z.-H. Zhou, ``Multi-view video
  summarization.'' \emph{IEEE Transactions on Multimedia}, vol.~12, no.~7, pp.
  717--729, 2010.

\bibitem{Leo:2014:MVS}
C.~d. Leo and B.~S. Manjunath, ``Multicamera video summarization and anomaly
  detection from activity motifs,'' \emph{ACM Transactions on Sensor Networks},
  vol.~10, no.~2, pp. 27:1--27:30, Jan. 2014.

\bibitem{conf/bmvc/VaradarajanEO10}
J.~Varadarajan, R.~Emonet, and J.-M. Odobez, ``Probabilistic latent sequential
  motifs: Discovering temporal activity patterns in video scenes.'' in
  \emph{British Machine Vision Conference}, 2010, pp. 1--11.

\bibitem{blei2003}
D.~M. Blei, A.~Y. Ng, and M.~I. Jordan, ``Latent dirichlet allocation,''
  \emph{Journal of Machine Learning Research}, vol.~3, pp. 993--1022, Mar.
  2003.

\bibitem{CLiu:BP}
C.~Liu, ``Beyond pixels: Exploring new representations and applications for
  motion analysis,'' Ph.D. dissertation, the Massachusetts Institute of
  Technology, 2009.

\bibitem{VaradarajanO_ICCV09}
J.~Varadarajan and J.~Odobez, ``Topic models for scene analysis and abnormality
  detection,'' in \emph{IEEE International Conference on Computer Vision,
  Computer Vision Workshops}, 2009, pp. 1338--1345.

\bibitem{fu2013latentAttrib}
Y.~Fu, T.~Hospedales, T.~Xiang, and S.~Gong, ``Learning multimodal latent
  attributes,'' \emph{IEEE Transactions on Pattern Analysis and Machine
  Intelligence}, vol.~36, no.~2, pp. 303--316, Feb 2014.

\bibitem{conf/nips/Zelnik-ManorP04}
L.~Zelnik-Manor and P.~Perona, ``Self-tuning spectral clustering.'' in
  \emph{Conference on Neural Information Processing Systems}, 2004.

\bibitem{DBLP:conf/iccv/ZhengJ13}
J.~Zheng and Z.~Jiang, ``Learning view-invariant sparse representations for
  cross-view action recognition,'' in \emph{IEEE International Conference on
  Computer Vision}, 2013, pp. 3176--3183.

\bibitem{ZhengJPC_BMVC12}
J.~Zheng, Z.~Jiang, P.~J. Phillips, and R.~Chellappa, ``Cross-view action
  recognition via a transferable dictionary pair,'' in \emph{British Machine
  Vision Conference}, 2012, pp. 1--11.

\bibitem{journals/tcs/Gonzalez85}
T.~F. Gonzalez, ``Clustering to minimize the maximum intercluster distance.''
  \emph{Theoretical Computer Science}, vol.~38, pp. 293--306, 1985.

\bibitem{Hochbaum1985}
Hochbaum and Shmoys, ``A best possible heuristic for the k-center problem,''
  \emph{Mathematics of Operations Research}, vol.~10, no.~2, pp. 180--184,
  1985.

\bibitem{Shi00normalizedCuts}
J.~Shi and J.~Malik, ``Normalized cuts and image segmentation,'' \emph{IEEE
  Trans. on Pattern Analysis and Machine Intelligence}, vol.~22, no.~8, pp.
  888--905, 2000.

\bibitem{journals/tcsv/NgoMZ05}
C.-W. Ngo, Y.-F. Ma, and H.~Zhang, ``Video summarization and scene detection by
  graph modeling.'' \emph{IEEE Trans. Circuits Syst. Video Techn.}, vol.~15,
  no.~2, pp. 296--305, 2005.

\bibitem{journals/tmm/MaHLZ05}
Y.-F. Ma, X.-S. Hua, L.~Lu, and H.~Zhang, ``A generic framework of user
  attention model and its application in video summarization.'' \emph{IEEE
  Transactions on Multimedia}, vol.~7, no.~5, pp. 907--919, 2005.

\bibitem{rand1971}
W.~Rand, ``Objective criteria for the evaluation of clustering methods,''
  \emph{Journal of the American Statistical Association}, vol.~66, no. 336, pp.
  846--850, 1971.

\bibitem{Shao_2014_CVPR}
J.~Shao, C.~C. Loy, and X.~Wang, ``Scene-independent group profiling in
  crowd,'' in \emph{IEEE Conference on Computer Vision and Pattern
  Recognition}, June 2014.

\end{thebibliography}
\end{document}